%% file: example_paper.tex
\documentclass{article}

\usepackage{microtype}
\usepackage{graphicx}
\usepackage{subcaption}
\usepackage{booktabs} % for professional tables
\usepackage{placeins}
\usepackage[colorlinks=true,linkcolor=red]{hyperref}

\usepackage[accepted]{icml2026}

\usepackage{amsmath}
\usepackage{amssymb}
\usepackage{mathtools}
\usepackage{amsthm}

\input{define}

\usepackage[capitalize,noabbrev]{cleveref}

\theoremstyle{plain}

\theoremstyle{definition}

\theoremstyle{remark}

\usepackage[textsize=tiny]{todonotes}

\usepackage{CJKutf8}

\begin{document}

\twocolumn[
  \icmltitle{
  Robustifying Vision-Language Models via Test-Time Prompt Adaptation}



  \icmlsetsymbol{equal}{*}

  \begin{icmlauthorlist}
    \icmlauthor{Xingyu Zhu}{nus,ustc}
    \icmlauthor{Huanshen Wu}{ustc}
    \icmlauthor{Shuo Wang}{ustc,equal}
    \icmlauthor{Beier Zhu}{ustc}
    \icmlauthor{Jiannan Ge}{ustc}
    \icmlauthor{Jiaheng Zhang}{nus}
    \icmlauthor{Long Chen}{hkust}

  \end{icmlauthorlist}

  \icmlaffiliation{ustc}{University of Science and Technology of China}
  % \icmlaffiliation{hw}{Huawei Technologies Ltd, Guangdong, China}
  \icmlaffiliation{nus}{National University of Singapore}
  \icmlaffiliation{hkust}{The Hong Kong University of Science and Technology}

  \icmlcorrespondingauthor{Shuo Wang}{shuowang.edu@gmail.com}

  \icmlkeywords{Machine Learning, ICML}

  \vskip 0.3in
]



\printAffiliationsAndNotice{}  % no special notice (required even if empty)

%Figure1的caption, (d)图的替换
%related work, method参考文献

\input{Sections/0_abstract}
\input{Sections/1_introduction}
\input{Sections/2_related_work}

\input{Sections/3_method}

\input{Sections/4_experiments}

\section*{Conclusion \& Limitations}
In this work, we propose $\ours$, a robust test-time prompt adaptation framework that enhances the adversarial robustness of VLMs without requiring retraining or access to labeled data. By modeling augmented visual features and prompt-induced textual prototypes as distributions and aligning them via optimal transport, $\ours$ corrects cross-modal semantic misalignment caused by adversarial perturbations. Furthermore, a dynamic cache mechanism progressively aggregates reliable semantic cues from the test stream to refine alignment online. Extensive experiments across diverse benchmarks, attack types, and model backbones demonstrate that $\ours$ consistently improves adversarial robustness while preserving competitive performance on clean data. Our results suggest that distribution-level alignment is a principled and effective paradigm for robust inference in large pre-trained VLMs.

While $\ours$ demonstrates strong image classification performance, its extension to generative tasks like image captioning remains for future exploration. We believe our distribution-level alignment provides a foundation for adaptation in these broader multimodal scenarios.

\section*{Impact Statement}
This work improves the reliability and adversarial robustness of pre-trained VLMs, which is important for their deployment in safety-sensitive applications. The proposed method operates entirely at test time, without modifying model parameters or requiring additional training data, making it lightweight and easy to integrate into existing systems. Nevertheless, it does not eliminate all security risks, and adversarial attacks may continue to evolve. Future work should further investigate robust inference-time defenses and evaluate potential failure modes and misuse risks.

\section*{Acknowledgement}
This research is supported by the National Natural Science Foundation of China (No. 62576330) and the National Natural Science Foundation of Anhui (No.2508085MF143).

\bibliography{example_paper}
\bibliographystyle{icml2026}

%%%%%%%%%%%%%%%%%%%%%%%%%%%%%%%%%%%%%%%%%%%%%%%%%%%%%%%%%%%%%%%%%%%%%%%%%%%%%%%
%%%%%%%%%%%%%%%%%%%%%%%%%%%%%%%%%%%%%%%%%%%%%%%%%%%%%%%%%%%%%%%%%%%%%%%%%%%%%%%
% APPENDIX
%%%%%%%%%%%%%%%%%%%%%%%%%%%%%%%%%%%%%%%%%%%%%%%%%%%%%%%%%%%%%%%%%%%%%%%%%%%%%%%
%%%%%%%%%%%%%%%%%%%%%%%%%%%%%%%%%%%%%%%%%%%%%%%%%%%%%%%%%%%%%%%%%%%%%%%%%%%%%%%
\newpage
\appendix
\onecolumn

\input{Sections/appendix}

%%%%%%%%%%%%%%%%%%%%%%%%%%%%%%%%%%%%%%%%%%%%%%%%%%%%%%%%%%%%%%%%%%%%%%%%%%%%%%%
%%%%%%%%%%%%%%%%%%%%%%%%%%%%%%%%%%%%%%%%%%%%%%%%%%%%%%%%%%%%%%%%%%%%%%%%%%%%%%%

\end{document}

%% file: define.tex
\def\eg{\emph{e.g.}} 

\def\ie{\emph{i.e.}}

\DeclareMathOperator*{\argmin}{argmin}
\DeclareMathOperator*{\argmax}{argmax}

\newcommand{\ours}{\texttt{RITA}}

\usepackage{amsthm}
\usepackage{thmtools} 
\usepackage{thm-restate}
\definecolor{lightCyan}{rgb}{0.925,1,1}
\definecolor{rowgreen}{rgb}{0.90,0.98,0.98}

\usepackage{multirow}
\usepackage{bbm}

\usepackage{xcolor,colortbl}

\definecolor{Gray}{gray}{0.9}
\newcolumntype{a}{>{\columncolor{Gray}}c}

\definecolor{blush}{rgb}{0.87, 0.36, 0.51}

\newcommand\bestclean[1]{\textbf{#1}}

\usepackage{bbding}
\usepackage{algorithm,algorithmic}

\newcommand{\tableCellHeight}{1}
\newcommand{\tabstyle}[1]{
  \setlength{\tabcolsep}{#1}
  \renewcommand{\arraystretch}{\tableCellHeight}
  \centering
  \small
}

%% file: Sections/0_abstract.tex
\begin{abstract}

Pre-trained Vision-Language Models (VLMs) such as CLIP achieve strong zero-shot generalization, but their performance degrades sharply under adversarial perturbations.
Existing test-time adaptation methods typically rely on sample-level confidence heuristics, overlooking the intrinsic distributional structure of the data.
This sample-centric approach limits robustness, as it fails to distinguish confident adversarial mispredictions from true semantic consistency.
In this work, we observe that adversarial distortion is structurally brittle: while holistic representations are corrupted, semantic integrity is often preserved in the distribution of augmented views.
Motivated by this insight, we propose $\ours$, 
a \textbf{R}obust test-t\textbf{I}me promp\textbf{T} \textbf{A}daptation framework that shifts from sample-level estimates to distribution-level alignment.
Specifically, $\ours$ employs optimal transport to align the distribution of augmented visual features with textual prototypes, mitigating adversarial outliers and rectifying cross-modal semantic misalignment.
Furthermore, we introduce a dynamic cache to progressively accumulate reliable cues from the test stream for online refinement.
Extensive experiments demonstrate that $\ours$ significantly improves adversarial robustness without compromising clean accuracy.

\end{abstract}
% xingyu zhu, huansheng wu, shuo wang, beier zhu, jianna ge, jiaheng zhang, long chen.

%% file: Sections/1_introduction.tex
\section{Introduction}

Vision-Language Models (VLMs)~\cite{BLIP, Flamingo,blip2, SelectiveXingyz, HierarchicalXingyz} like CLIP~\cite{RadfordCLIP21}, pre-trained on massive image-text pairs, have achieved remarkable zero-shot generalization. Despite this success, VLMs remain highly vulnerable to adversarial perturbations: imperceptible noise can cause severe performance degradation~\cite{SzegedyZSBEGF13,MadryMSTV18, EnhancingXingyz, COLA}, posing security risks in real word applications.

\input{Figures/motivation}
Existing efforts to enhance the robustness of VLMs generally fall into two categories.
The first line of work utilizes \emph{adversarial training}~\cite{TeCoA, FARE, PMG-AFT, advPT}, which aims to immunize models by explicitly integrating adversarial examples into the optimization loop.
While effective, these approaches typically incur prohibitive computational costs due to on-the-fly attack generation and require access to task-specific labeled data, thereby undermining the scalability and zero-shot flexibility inherent to foundation models.
The second line of work explores \emph{test-time prompt tuning}~\cite{C-TPT, TPT, abs-2505-17509}, an efficient paradigm that adapts learnable prompt contexts or predictions during inference without modifying model parameters.
However, most existing test-time methods~\cite{TAPT, RTPT} primarily rely on {sample-level} confidence (\eg, entropy) to filter augmented views. 
These methods treat augmentations as isolated data points, overlooking the intrinsic {distributional structure} and latent semantics.
Consequently, they fail to distinguish between confident adversarial {mispredictions} and true semantic consistency, limiting their effectiveness under attack.

To address the above limitation, we revisit how adversarial perturbations interact with test-time augmentations in VLMs.
A typical white-box attack is crafted on the original  input image to maximally disrupt image-text matching based on its holistic representation.
This can drastically shift the attacked image embedding in the feature space and make samples from different classes highly entangle as shown in Figure~\ref{fig:motivation}(a).
Importantly, this adversarial effect is not equally persistent across augmentations.
Geometric transformations such as random cropping and flipping change the spatial correspondence of pixels, so the perturbation pattern optimized for the original configuration becomes partially mismatched after transformation.
Empirically, different views exhibit heterogeneous behaviors: while some views remain strongly influenced by the attack, others still produce confident and semantically consistent predictions (Figure~\ref{fig:motivation}(b)).
This observation suggests that robustness should be built by exploiting the distribution of augmented views and the semantic relations it contains, rather than relying solely on sample-level confidence heuristics.

Motivated by the above observation, we propose $\ours$, a robust test-time framework that shifts from sample-level matching to {distribution-level alignment}.
Instead of relying on a single image embedding, 
$\ours$ models the visual input as a discrete distribution over augmented views. 
To bridge these visual features with textual prototypes, we formulate the alignment objective as an {Optimal Transport (OT)}~\cite{OT} problem. 
As illustrated in Figure~\ref{fig:motivation}(c), this formulation enables us to evaluate the global geometric correspondence between the visual distribution and textual representations,  mitigating the influence of adversarial outliers to {rectify} semantic misalignment.
Moreover, test samples arrive as a continuous stream, providing additional information beyond a single image. To leverage this property, $\ours$ incorporates a {dynamic cache mechanism} that progressively accumulates reliable semantic views, and uses them to further refine distribution alignment online. As demonstrated in Figure~\ref{fig:motivation}(d), this progressive adaptation significantly enhances zero-shot robustness and maintaining competitive performance on clean data.

Extensive experiments on multiple standard benchmarks under diverse adversarial attacks demonstrate that $\ours$ significantly enhances zero-shot robustness while preserving competitive performance on clean data.
Our contributions are summarized as follows:
\begin{itemize}
    % Point 1: RITA + Augmented Views (The Insight/Source)
    \item We propose $\ours$, a robust test-time prompt adaptation framework that leverages \textbf{augmented views} to rectify adversarial misalignment at the distribution level.
    
    \item We formulate cross-modal semantic alignment as an  Optimal Transport problem, complemented by a dynamic cache for {progressive refinement}.
    
    % Point 3: Experiments
    \item We conduct comprehensive evaluations, showing that $\ours$ consistently outperforms existing test-time adaptation methods.
\end{itemize}

%% file: Figures/motivation.tex
\begin{figure}[!t]
  \centering
\includegraphics[width=1\linewidth]{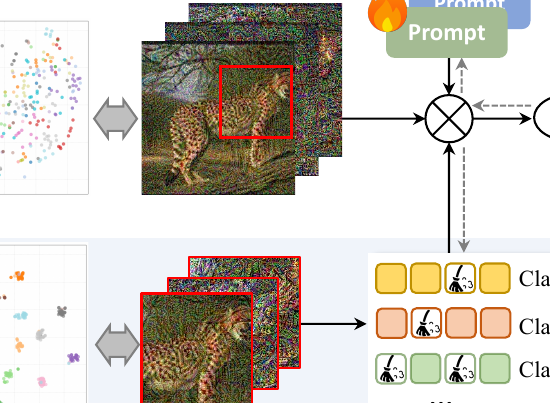}
    \caption{
% We visualize adversarial images and their augmented views, leverage selected views as a cache for alignment, and compare performance across different VLM backbones.
% (a) Visualization results of adversarially perturbed images in the visual embedding space.
% (b) Visualization results of multiple augmented views from the same adversarial image, showing improved separability compared to (a).
% (c) Our method leverages the selected augmented views as a cache and aligns them with textual prompts.
% (d) Performance comparison across different VLM backbones, demonstrating improved robustness under adversarial attacks.
Augmented views retain more semantic cues under adversarial perturbations, enabeling a cache for distribution alignment that improves adversarial performance.
(a) Visualization of adversarially perturbed images, where each point represents an image and different colors denote ground-truth classes.
(b) Visualization of multiple augmented views generated from the same adversarial image, colored by class label, where semantic structure partially re-emerges with improved class separability compared to (a).
(c) Our method leverages the selected augmented views as a cache and aligns them with textual prompts.
(d) Performance comparison across different VLM backbones, demonstrating improved robustness under adversarial attacks.
}\label{fig:motivation}
\end{figure}

%% file: Sections/2_related_work.tex
\section{Related Work}

\noindent{\textbf{Adversarial Defense in VLMs.}}
% The vulnerability of VLMs (\eg, CLIP~\cite{RadfordCLIP21}) to adversarial perturbations remains a critical challenge~\cite{DongLPS0HL18,MadryMSTV18,ZhaoPDYLCL23}. Attacks have evolved from uni-modal perturbations~\cite{carlini2017towards,atk-DI} to multi-modal strategies like Co-Attack~\cite{Co-Attack} that disrupt cross-modal alignment.
% To enhance robustness, prior defenses utilize training-time strategies, primarily adversarial contrastive tuning~\cite{TeCoA, FARE, PMG-AFT}. However, these methods typically require expensive re-training and labeled data, limiting their open-world practicality.
% Consequently, inference-time robustness has emerged to secure models without weight updates, such as diffusion purification~\cite{DiffTPT, FengHZKXLZGL25} and optimization-based methods~\cite{TENT, memo}.
% Notably, Test-Time Prompt Tuning approaches like TAPT~\cite{TAPT} and R-TPT~\cite{RTPT} adapt prompts using unlabeled data via contrastive learning or entropy minimization.
% Nevertheless, these paradigms typically treat visual inputs as isolated points, failing to exploit the underlying distributional geometry of adversarial samples.
% In contrast, we propose $\ours$, a test-time prompt tuning framework that shifts from point-level alignment to distribution-level modeling.
The vulnerability of VLMs (\eg, CLIP~\cite{RadfordCLIP21}) to adversarial perturbations remains a critical challenge~\cite{DongLPS0HL18,MadryMSTV18,ZhaoPDYLCL23, NullSteer}. Attacks have evolved from uni-modal perturbations~\cite{carlini2017towards,atk-DI} to multi-modal strategies like Co-Attack~\cite{Co-Attack} that disrupt cross-modal alignment.
To enhance robustness, prior defenses utilize training-time strategies, primarily adversarial contrastive tuning~\cite{TeCoA, FARE, PMG-AFT}. However, these methods typically require expensive re-training and labeled data, limiting practicality.
Consequently, inference-time robustness has emerged to secure models without weight updates, such as diffusion purification~\cite{DiffTPT, FengHZKXLZGL25} and optimization-based methods~\cite{TENT, memo}.
Notably, Test-Time Prompt Tuning approaches like TAPT~\cite{TAPT} and R-TPT~\cite{RTPT} adapt prompts using unlabeled data via contrastive learning or entropy minimization.
Nevertheless, these paradigms typically treat visual inputs as isolated points, failing to exploit the underlying distributional geometry of adversarial samples.
In contrast, we propose $\ours$, a test-time prompt tuning framework that shifts from point-level alignment to distribution-level modeling.

\noindent{\textbf{Optimal transport.}}
Optimal Transport (OT)~\cite{OT} provides a principled way to compare probability distributions by accounting for the geometry of the underlying feature space. With efficient solvers such as Sinkhorn~\cite{AltschulerWR17,MenschP20}, OT has been widely used in generative modeling~\cite{arjovsky2017wasserstein}, structural alignment~\cite{xu2019scalable}, and domain adaptation~\cite{courty2017optimal}. In vision--language learning, OT has also been applied to reduce semantic misalignment, including few-shot learning~\cite{iLPC}, distribution calibration~\cite{GuoT0ZZ22, DamodaranKFTC18, ProtoMM, COLA}, and prompt learning~\cite{plot, align, RenTZZGC25}. For example, PLOT~\cite{plot} aligns image features with multiple prompts via OT-based matching to capture diverse semantics, while ALIGN~\cite{align} further introduces hierarchical/token-level transportation for fine-grained cross-modal alignment~\cite{LookXingyz, GuardXingyz}. AWT~\cite{AWT} similarly formulates image-text distance as an OT problem to model semantic correlations in the joint space. However, these approaches primarily focus on representation enhancement in clean settings. They rely on undistorted visual manifolds and do not account for the severe structural perturbations caused by adversarial attacks. Differing from these approaches, $\ours$ repurposes OT for adversarial defense, aiming to reconstruct the distribution-level correspondence disrupted by attacks. This enables inference-time correction of structural misalignment without parameter updates, ensuring robust deployment.

%% file: Sections/3_method.tex
\section{Method}
% In this section, we detail the proposed RITA method.
% We begin with necessary preliminaries in Section~\ref{sec:method_preliminary}, followed by the method in Sections~\ref{sec:distribution} and~\ref{sec:alignment}, and conclude with theoretical analysis in Section~\ref{sec:theory}.
% Figure~\ref{fig:framework} illustrates the overall pipeline.

An overview of $\ours$ is illustrated in Figure~\ref{fig:framework}. We first introduce the preliminaries in Sec.~\ref{sec:method_preliminary}, then detail our distributed feature modeling and dynamic distribution alignment in Sec.~\ref{sec:distribution} and~\ref{sec:alignment}, respectively. Finally, we provide a theoretical justification in Sec.~\ref{sec:theory}.
\subsection{Preliminary}\label{sec:method_preliminary}
\noindent\textbf{Test-time prompt tuning.}
Test-Time Prompt Tuning (TPT)~\cite{TPT} improves the zero-shot generalization of CLIP~\cite{RadfordCLIP21} by adapting textual prompts at inference time, without accessing labeled data or updating CLIP parameters.
Given a test image $x_t$, TPT constructs a set of $N$ stochastic augmentations
$\mathcal{X}_t=\{x_t^{n}\}_{n=1}^{N}$.
For each augmented view $x_t^{n}$, the visual representation is $
\mathbf{x}_t^{n}=\Phi_{\mathrm{img}}(x_t^{n}),
$
where $\Phi_{\mathrm{img}}(\cdot)$ denotes the CLIP image encoder.
On the text side, the prompt for class $k$ is formulated as
$
z_k=\{\omega_1,\omega_2,\dots,\omega_L,c_k\},
$
where $c_k$ is the token embedding of the class name, and
$\omega=\{\omega_\ell\}_{\ell=1}^{L}$ are learnable context vectors shared across classes.
The textual representation is
$
\mathbf{z}_k=\Phi_{\mathrm{text}}(z_k),
$
with $\Phi_{\mathrm{text}}(\cdot)$ being the CLIP text encoder.
For each view $x_t^n$, the prediction probability is computed by feature matching:
\begin{equation}\label{eq:zero_shot}
p(k \mid x_t^n; \omega)
= \frac{\exp \big( \cos(\mathbf{x}_t^n, \mathbf{z}_k) / \tau \big)}
{\sum_{j=1}^K \exp \big( \cos( \mathbf{x}_t^n, \mathbf{z}_j ) / \tau \big)},
\end{equation}
where $\tau$ is the temperature parameter and $K$ is the number of categories. 
Specifically, TPT employs a confidence selection strategy to filter out unreliable augmentations. 
It selects a subset $\mathcal{S}$ of low-entropy views to compute the aggregated prediction $\bar{p}(k \mid x_t; \omega) = \frac{1}{|\mathcal{S}|} \sum_{n \in \mathcal{S}} p(k \mid x_t^n; \omega)$. 
The prompt $\omega$ is then optimized by minimizing the entropy of this aggregated distribution $\bar{p}$:
\begin{equation}
\omega^{*}
= \operatorname*{argmin}_{\omega}
\Big(
-\sum_{k=1}^K \bar{p}(k \mid x_t; \omega)
\log \bar{p}(k \mid x_t; \omega)
\Big).
\end{equation}

\input{Figures/framework}

\noindent{\textbf{Optimal transport.}}
Optimal Transport (OT)~\cite{OT} provides a principled way to measure the discrepancy between two probability distributions.
Consider two discrete distributions in the feature space,
$\mathbbm{P} = \sum_{n=1}^{N} a^n \delta_{\mathbf{x}^n}$ and
$\mathbbm{Q} = \sum_{m=1}^{M} b^m \delta_{\mathbf{z}^m}$,
where $\delta_{\mathbf{v}}$ denotes the Dirac delta function at location $\mathbf{v}$, and $\mathbf{a} \in \Delta_N, \mathbf{b} \in \Delta_M$ are probability vectors.
Given a cost matrix $\mathbf{C} \in \mathbb{R}^{N \times M}$, where $\mathbf{C}_{nm}$ measures the transport cost from $\mathbf{x}^n$ to $\mathbf{z}^m$,
the entropy-regularized OT distance is defined as:
\begin{equation}
d_{\mathrm{OT}}(\mathbbm{P},\mathbbm{Q};\mathbf{C})
= \min_{\mathbf{T} \in \Pi(\mathbf{a}, \mathbf{b})}
\langle \mathbf{T}, \mathbf{C} \rangle
- \lambda h(\mathbf{T}),
\end{equation}
where $\Pi(\mathbf{a}, \mathbf{b}) = \{ \mathbf{T} \in \mathbb{R}_+^{N \times M} \mid \mathbf{T}\mathbbm{1}_{M}=\mathbf{a}, \mathbf{T}^{\top}\mathbbm{1}_{N}=\mathbf{b} \}$ is the transport polytope, $h(\mathbf{T}) = -\sum_{n,m} T_{nm}\log T_{nm}$ is the entropic regularization term, and $\lambda \geq 0$ controls the regularization strength. This formulation enables efficient computation via the Sinkhorn algorithm.

\subsection{Distributed Features Modeling}\label{sec:distribution}
In this section, we present our core strategy for robust test-time inference. Moving beyond vulnerable {holistic} image embeddings, we model the adversarial image and textual prompts as discrete distributions and align them structurally using optimal transport.
% \noindent{\textbf{Adversarial Perturbations.}} 
% Adversarial attacks aim to degrade model predictions by introducing small, carefully crafted perturbations. In a white-box setting, given a clean image $x$ and ground-truth label $y_k$, the adversarial example $x'$ is generated by maximizing a task-specific loss under a norm-bounded constraint:
% \begin{equation}
%     x' = \operatorname*{argmax}_{\|x' - x\|_p \leq \epsilon_{\mathrm{adv}}}
% \mathcal{L}_{\mathrm{CE}}\big(p(y \mid x'; \{c_k\}_{k=1}^K), y_k\big),
% \end{equation}
% where $\epsilon_{\mathrm{adv}}$ denotes the perturbation budget and $p(\cdot)$ represents the CLIP zero-shot probability. We approximate this optimization using Projected Gradient Descent (PGD) to disrupt the alignment between image and text representations.

\noindent{\textbf{Adversarial perturbations.}} 
Adversarial attacks aim to degrade model predictions by introducing small, carefully crafted perturbations. In a white-box setting, given a clean image $x$ and its ground-truth label $y$, the adversarial example $x'$ is generated by maximizing the cross-entropy loss within an $\ell_p$-norm constraint:
\begin{equation}
    x' = \operatorname*{argmax}_{\|x' - x\|_p \leq \epsilon_{\mathrm{adv}}}
    \mathcal{L}_{\mathrm{CE}}\Big(p(\cdot \mid x'; z), y\Big),
\end{equation}
where $\epsilon_{\mathrm{adv}}$ denotes the perturbation budget, and $p(\cdot \mid x'; z)$ represents the probability distribution over $K$ classes as defined in Eq.~\eqref{eq:zero_shot}. Typically,  we approximate this optimization using the iterative Projected Gradient Descent (PGD)~\cite{MadryMSTV18}.

\noindent{\textbf{Multi-prototype distribution alignment.}} 
While adversarial attacks distort alignment at the global representation level, relying on single-point embeddings is insufficient to capture the semantic variations. To alleviate this, we construct sets of diverse visual and textual representations and model their correspondence at the distribution level.

Specifically, given an adversarial image $\hat{x}_t$, we apply data augmentations to obtain $N$ views $\{\hat{x}_t^n\}_{n=1}^{N}$, producing visual features $\{\hat{\mathbf{x}}_t^n\}_{n=1}^{N}$. 
Instead of a single text prototype, for each class $k$, we construct $M$ prompts $\{z_k^{(m)}\}_{m=1}^{M}$ using learnable context vectors, yielding textual features $\{\mathbf{z}_k^{m}\}_{m=1}^{M}$.
We model the adversarial image and the $k$-th class prototype as discrete distributions:
\begin{equation}
  \mathbbm{P}_{t} = \sum_{n=1}^{N} \frac{1}{N} \, \delta_{\hat{\mathbf{x}}_t^n}, \quad
  \mathbbm{Q}_{k} = \sum_{m=1}^{M} \frac{1}{M} \, \delta_{\mathbf{z}_k^{m}}.
\end{equation}

Here, we assign uniform weights to both modalities, \ie, $\mathbf{a}_t=\frac{1}{N}\mathbbm{1}_N$ and $\mathbf{b}_k=\frac{1}{M}\mathbbm{1}_M$, such that the marginal constraints of OT are satisfied. 
We then measure the distribution-level alignment between the visual distribution $\mathbbm{P}_t$ and the textual prototype distribution $\mathbbm{Q}_k$ via entropy-regularized optimal transport:
\begin{equation}
d_{\mathrm{OT}}(\mathbbm{P}_t, \mathbbm{Q}_k; \mathbf{C}_{t,k})
=
\min_{\mathbf{T}_{t,k} \in \Pi}
\left(
\langle \mathbf{T}_{t,k}, \mathbf{C}_{t,k} \rangle -
\lambda 
h(\mathbf{T}_{t,k})
\right),
\end{equation}
where $\mathbf{T}_{t,k}\in\mathbb{R}_+^{N\times M}$ is the transport plan, and the cost matrix is defined by the cosine distance,
$\mathbf{C}_{t,k}(n,m)=1-\cos(\hat{\mathbf{x}}_t^{\,n}, \mathbf{z}_k^{\,m})$.
Intuitively, a smaller OT distance indicates better alignment between the test image and class $k$ at the distribution level.
Accordingly, we predict the label by selecting the class with the minimum transport cost:
\begin{equation}
\hat{y}
=
\operatorname*{argmin}_{k\in[K]}
d_{\mathrm{OT}}(\mathbbm{P}_t, \mathbbm{Q}_k; \mathbf{C}_{t,k}).
\end{equation}

\subsection{Dynamic Distribution Alignment}\label{sec:alignment}
% Following the observation that low-entropy predictions provide reliable semantic cues, we select confident augmented views to build a dynamic cache for distribution alignment.
While distribution-level alignment in Sec.~\ref{sec:distribution} effectively processes individual samples, it treats inference steps in isolation, neglecting the semantic consensus in the continuous test stream. 
To exploit this temporal information, we introduce a {dynamic cache mechanism} that accumulates reliable visual features to iteratively refine the alignment online.

\noindent\textbf{Confidence-based cache update.}
To update the cache with reliable samples, we evaluate the confidence of each individual augmented view. Instead of solving the full global transport problem, we quantify the instance-level alignment by measuring the average affinity between the view $\hat{\mathbf{x}}_t^n$ and the text distribution of class $k$:
\begin{equation}
p_t^{n}(k)
=
\frac{\exp\!\left( \frac{1}{M}\sum_{m=1}^{\tau M}\cos(\hat{\mathbf{x}}_t^{\,n},\mathbf{z}_k^{m})  \right)}
{\sum_{j=1}^{K}\exp\!\left(\frac{1}{M}\sum_{m=1}^{\tau M}\cos(\hat{\mathbf{x}}_t^{\,n},\mathbf{z}_j^{m})\right)}.
\end{equation}
We compute entropy $H(p_t^{n})=-\sum_{k=1}^{K}p_t^{n}(k)\log p_t^{n}(k)$ and retain confident views:
\begin{equation}    
\mathcal{B}_t=\left\{\, (\hat{\mathbf{x}}_t^{\,n},\hat{y}_t^{\,n})\ \middle|\ H(p_t^{n})\le \gamma,\ 
\hat{y}_t^{\,n}=\operatorname*{argmax}_k p_t^{n}(k)\,\right\}.
\end{equation}
% We compute the entropy as $H(p_t^{n}) = -\sum_{k=1}^{K} p_t^{n}(k) \log p_t^{n}(k)$ for each augmented view. To filter for the most reliable predictions, we sort all $N$ views by entropy in ascending order and retain the top $\lfloor p \cdot N \rfloor$ samples (i.e., those with the lowest entropy) to form the confident subset $\mathcal{B}_t$:
% \begin{equation}
% \mathcal{B}_t = \{ (\hat{\mathbf{x}}_t^{\,n}, \hat{y}_t^{\,n}) \mid \text{rank}(H(p_t^{n})) \le \lfloor p \cdot N \rfloor \},
% \end{equation}
% where $\hat{y}_t^{\,n} = \operatorname*{argmax}_k p_t^{n}(k)$ is the predicted label for each selected view, $p \in (0, 1]$ is the confidence selection ratio and $\text{rank}(\cdot)$ denotes the index in the sorted sequence.
We maintain a class-wise cache $\{\hat{\mathbf{X}}_k\}_{k=1}^{K}$, where
$\hat{\mathbf{X}}_k\in\mathbb{R}^{N_k\times d}$ stores cached visual features pseudo-labeled as class $k$
($N_k$ is the current cache size). The cache is updated online by prioritizing lower-entropy samples.
To bridge the modality gap, we align cached visual features to the textual space of class $k$ by solving an Orthogonal Procrustes~\cite{LFA} problem:
\begin{equation}
\mathbf{W}_k^\ast
=
\operatorname*{argmin}_{\mathbf{W}^\top \mathbf{W}=\mathbf{I}}
\left\|
\hat{\mathbf{X}}_k \mathbf{W}
-
\mathbf{1}_{N_k}\bar{\mathbf{z}}_k^\top
\right\|_F^2,
\end{equation}
where $\bar{\mathbf{z}}_k = \frac{1}{M}\sum_{m=1}^M \mathbf{z}_k^m$ is the mean text embedding for class $k$.
The aligned cached features are $\tilde{\mathbf{X}}_k = \hat{\mathbf{X}}_k \mathbf{W}_k^\ast$.

\noindent\textbf{Cache-based distribution matching.}
For each class $k$, we instantiate a discrete probability distribution over the aligned cached features:
\begin{equation}
\tilde{\mathbbm{Q}}_k
=
\frac{1}{N_k}\sum_{j=1}^{N_k}\delta_{\tilde{\mathbf{x}}_k^{\,j}},
\qquad \text{where } \tilde{\mathbf{x}}_k^{\,j}\in \tilde{\mathbf{X}}_k.
\end{equation}
We quantify the discrepancy via the same entropy-regularized OT distance as in Sec.~\ref{sec:distribution}, but with a cache-specific cost matrix $\tilde{\mathbf{C}}_{t,k}(n,j)=1-\cos(\hat{\mathbf{x}}_t^{\,n},\tilde{\mathbf{x}}_k^{\,j})$, and denote it as $d_{\mathrm{cOT}}(\mathbbm{P}_t, \tilde{\mathbbm{Q}}_k; \tilde{\mathbf{C}}_{t,k})$.

\noindent\textbf{Final inference.}
We classify the test sample by identifying the category that minimizes the joint transport cost, which integrates both the global prompt alignment and the local cache consensus:
\begin{equation}
\label{eq:final_pred}
\hat{y}
=
\operatorname*{argmin}_{k\in[K]}
\Big(
d_{\mathrm{OT}}(\mathbbm{P}_t, \mathbbm{Q}_k; \mathbf{C}_{t,k})
+\alpha\, d_{\mathrm{cOT}}(\mathbbm{P}_t, \tilde{\mathbbm{Q}}_k; \tilde{\mathbf{C}}_{t,k})
\Big),
\end{equation}
where $\alpha\ge 0$ controls  the contribution of the dynamic cache.

\subsection{Theoretical Analysis}
\label{sec:theory}

Standard methods like TPT~\cite{TPT} and R-TPT~\cite{RTPT} optimize centroid alignment via mean pooling and cosine similarity. For $\ell_2$-normalized CLIP features, this is equivalent to minimizing the squared Euclidean distance between centroids ($\Vert\mathbf{x}-\mathbf{z}\Vert^2 = 2(1-\cos(\mathbf{x},\mathbf{z}))$). 
To reveal $\ours$'s geometric advantage, we analyze alignment using the 2-Wasserstein distance ($W_2$).
Let $\mathbb{P}_t$ ($\boldsymbol{\mu}_x,\boldsymbol{\Sigma}_x$) and $\mathbb{Q}_k$ ($\boldsymbol{\mu}_z,\boldsymbol{\Sigma}_z$) be the visual and textual distributions. We obtain the following decomposition:

% \noindent\textbf{Theorem 1 (Decomposition of Alignment Objective).} 
\begin{restatable}[Decomposition of Alignment Objective]{theorem}{decomAlign}
\textit{The Optimal Transport objective ($\mathcal{L}_{\mathrm{OT}}$) imposes a stricter bound by decomposing into a centroid alignment term and a structural variance penalty:}
\begin{equation}
\label{eq:ot_decomposition}
\underbrace{W_2^2(\mathbb{P}_t, \mathbb{Q}_k)}_{\mathcal{L}_{\mathrm{OT}} \text{ (RITA)}} 
\approx
\underbrace{\| \boldsymbol{\mu}_x - \boldsymbol{\mu}_z \|^2}_{\mathcal{L}_{\mathrm{mean}} \text{ (TPT-equivalent)}} 
+ 
\underbrace{\mathfrak{B}^2(\boldsymbol{\Sigma}_x, \boldsymbol{\Sigma}_z)}_{\mathcal{R}_{\mathrm{var}} \text{ (Structural Penalty)}},
\end{equation}
\textit{where $\mathfrak{B}^2(\mathbf{A}, \mathbf{B}) = \operatorname{Tr}(\mathbf{A} + \mathbf{B} - 2(\mathbf{A}^{1/2}\mathbf{B}\mathbf{A}^{1/2})^{1/2})$ is the Bures metric, quantifying the geometric mismatch between covariances.}
\end{restatable}
Equation~\eqref{eq:ot_decomposition} holds exactly for Gaussian distributions and serves as a general lower bound, as derived in {Appendix~\ref{proof}}. 
This inequality highlights a critical robustness gap: minimizing only $\mathcal{L}_{\mathrm{mean}}$ leaves structural variance $\mathcal{R}_{\mathrm{var}}$ unconstrained, allowing attackers to distort distribution geometry.

%% file: Figures/framework.tex
\begin{figure*}[!ht]
  \centering
\includegraphics[width=1\linewidth]{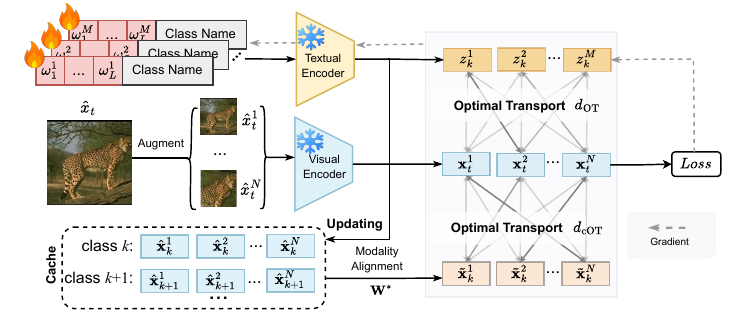}
    \caption{Overview of the proposed RITA framework.
Given an adversarial test image, RITA extracts multi-view visual features and class-specific textual prototypes using a frozen CLIP encoder. Both modalities are modeled as discrete distributions and aligned via entropy-regularized optimal transport. Low-entropy views are used to update a dynamic cache of reliable semantics.}
    \label{fig:framework}
\end{figure*}

%% file: Sections/4_experiments.tex
\section{Experiments}
\subsection{Setup}
\noindent{\textbf{Datasets.}} We evaluate our method on eight image classification benchmarks spanning a wide range of visual domains, including generic object recognition (Caltech101~\cite{Caltech}), texture recognition (DTD~\cite{DTD}), satellite imagery (EuroSAT~\cite{Eurosat}), human action recognition (UCF101~\cite{UCF101}), as well as several fine-grained classification tasks, namely Pets~\cite{OxfordPet}, Cars~\cite{Cars}, Flowers~\cite{Flower}, and Aircraft~\cite{FGVC}.
To further assess robustness under distribution shifts, we conduct additional evaluations on ImageNet~\cite{Imagenet} and four of its variants that share the same label space: ImageNetV2~\cite{ImageNetV2}, ImageNet-Sketch~\cite{ImageNetSketch}, ImageNet-A~\cite{ImageNetA}, and ImageNet-R~\cite{ImageNetR}. These benchmarks introduce significant variations in image sources, styles, and underlying visual statistics. Detailed analysis of these datasets is provided in Appendix~\ref{app_Imagenet}.

% All experiments are performed in a test-time adaptation setting, where neither training data nor ground-truth labels are accessible. Model adaptation relies exclusively on unlabeled test samples.

\noindent\textbf{Implementation details.}  
We build all experiments on the official pre-trained CLIP models with two backbones, CLIP-ViT-B/32 and CLIP-ViT-B/16. We generate adversarial examples using PGD~\cite{MadryMSTV18} under an $L_\infty$ constraint. We use $\epsilon=4.0$ with 7 steps for both backbones.
At test time, we update only the prompt parameters while keeping the CLIP backbone freezed. The prompt number $M$ is set to 4 and initialized with the template \emph{``a photo of a''}. We use AdamW and fix the Test-Time Adaptation (TTA) step at 1 per test sample, with a learning rate of 0.005.
We apply standard test-time augmentations for images, including random cropping, resizing, and horizontal flipping. For text, we use an Large Language Model (LLM) to generate class-specific descriptions\cite{AWT}. For each test image, we sample $N=64$ augmented views (including the original). For cache construction, We set the entropy threshold to $\gamma=0.8$. More experiment details are provided in Appendix~\ref{app:exp_detail}.

\noindent{\textbf{Comparison methods.}}
We compare \texttt{RITA} with CLIP-based test-time adaptation baselines, including TPT~\cite{TPT}, R-TPT~\cite{RTPT}, C-TPT~\cite{C-TPT}, and MTA~\cite{MTA}, as well as the zero-shot CLIP baseline. We also report an \textbf{Ensemble} baseline that averages predictions over augmented views. To further evaluate compatibility with robust pre-trained weights, We incorporate three representative adversarially fine-tuned CLIP models, namely TeCoA~\cite{TeCoA}, PMG~\cite{PMG-AFT}, and FARE~\cite{FARE}.
All methods follow the instance-level test-time adaptation protocol: each test sample is adapted and predicted independently, without access to other test samples.

\input{Tables/res_vitb32-16}

\input{Tables/ablation_finetuning}
\subsection{Main Results}
\noindent{\textbf{Results on fine-grained datasets.}}
%如表1所示，我们在两种架构（ViT-B/32和ViT-B/16）上对8个细粒度基准数据集进行了广泛测试。实验结果表明，RITA在绝大多数数据集的对抗扰动下均取得了最优精度。具体而言，与原始CLIP相比，RITA在ViT-B/32和ViT-B/16架构上的平均对抗鲁棒性分别大幅提升了 45.0% 和 50.9%，彻底改变了模型在面对攻击时几近失效（例如在Cars和Aircraft上精度接近0%）的局面。此外，与当前最先进的（SOTA）测试时自适应方法 R-TPT 相比，RITA 在两种架构下的平均鲁棒精度也分别提升了 1,8% 和 2.2%。此外，RITA 在干净样本上的平均精度也达到最优。这有力地证明了 RITA 在显著提升模型对抗鲁棒性的同时，能够极好地保持模型在无攻击环境下的原始识别性能。值得注意的是，RITA在ViT-B/16上的表现显著优于ViT-B/32，这与“更精细的ViT骨干网络具备更强识别能力”的直觉相符。
We evaluate our method on eight fine-grained benchmark datasets with ViT-B/32 and ViT-B/16 backbones, as presented in Table~\ref{tab:res_vitb32-16}. The results show that $\ours$ achieves the highest adversarial accuracy on nearly all datasets. Compared to vanilla CLIP, $\ours$ improves average robustness by 45.0\% and 50.9\% with ViT-B/32 and ViT-B/16, respectively. This effectively mitigates the severe vulnerability of the baseline under adversarial attacks, which can cause near-zero accuracy on datasets such as Cars and Aircraft. Furthermore, compared to R-TPT, the state-of-the-art test-time adaptation method, $\ours$ achieves consistent gains in average robust accuracy, outperforming it by 1.8\% and 2.2\% on the two architectures. Meanwhile, our method also achieves the highest average accuracy on clean samples.This demonstrates that $\ours$ enhances adversarial robustness while preserving recognition performance in attack-free environments with negligible degradation. Notably, $\ours$ performs better with ViT-B/16 than with ViT-B/32, aligning with the intuition that a fine-grained ViT backbone yields stronger recognition capabilities.

\noindent{\textbf{Results with adversarially finetuned CLIP models.}}
%TeCoA, PMG, FARE这三个backbone, 参考CLIP is Strong Enough to Fight Back: Test-time Counterattacks towardsZero-shot Adversarial Robustness of CLIP中的凯源代码、都有提供
%因为我们的方法天然具备可拔插性，我们将其外接在三种代表性对抗微调的CLIP模型（TeCoA~\cite{TeCoA}、PMG~\cite{PMG-AFT} 和 FARE~\cite{FARE}）中，结果见表~\ref{tab:ablation_finetuning}。我们发现，$\ours$ 能够与这些预训练的鲁棒模型产生显著的正向协同效应。在所有三种基准模型上，$\ours$ 均显著提升了各细粒度数据集的准确率与鲁棒精度。值得注意的是，在 Caltech101 数据集上，相比于基础的 FARE 模型，结合$\ours$ 后鲁棒精度从 62.9% 大幅提升至 81.9%。这有力地证明了我们的方法并非仅能提升原始 CLIP 的性能，它作为一种通用的测试时自适应框架，可以无缝集成到现有的对抗防御模型中，并进一步强化其在现实推理场景下的稳健性。
%由于 $\ours$ 具有天然的即插即用特性，我们将其与三种代表性的对抗微调模型（TeCoA、PMG 和 FARE）进行了集成。如表~\ref{tab:ablation_finetuning} 所示，$\ours$ 与这些鲁棒基线模型表现出显著的协同效应，在不损失清洁精度的前提下，大幅提升了各数据集的鲁棒准确率。值得注意的是，与 FARE 模型结合时，$\ours$ 在 Caltech101 上的鲁棒精度达到了惊人的 81.9%，较基线提升了 19.0%。这些结果证明 $\ours$ 是一种通用的自适应框架，能够无缝增强现有对抗防御模型在推理阶段的稳定性。
As an inherently plug-and-play framework, we integrated $\ours$ with three representative adversarially fine-tuned models, with results summarized in Table~\ref{tab:ablation_finetuning}. $\ours$ exhibits significant synergy with these robust baselines, substantially enhancing robust accuracy across all datasets without sacrificing clean performance. Notably, when combined with FARE, $\ours$ achieves a striking 81.9\% robust accuracy on Caltech101, representing a 19.0\% improvement over the baseline. These results demonstrate that $\ours$ is a versatile adaptation framework that can seamlessly fortify existing adversarial defense models during inference.

\noindent{\textbf{Results under different attack types.}}
\input{Tables/res_attack_type}
%为了进一步验证 RITA 的通用性，我们在表3中展示了模型在 CW 和 DI 两种攻击方式下的表现。实验结果显示，RITA 在不同攻击协议下均展现出了卓越的防御能力。具体而言，在 CW 攻击下，RITA 在 Pets、Flower102、Aircraft、DTD 等多个数据集上均取得了最优的对抗鲁棒性，在8个细粒度数据集的平均对抗精度达到了 54.0%，显著优于其余的 baseline 方法。在更具挑战性的 DI 攻击下，RITA 依然保持了领先地位，平均精度达到 46.2%，相比 R-TPT 提升了 2.0%。这一结果有力证明了 RITA 并非过拟合于某种特定的扰动模式，而是通过有效的测试时自适应机制，增强了预训练视觉语言模型在多种对抗威胁下的本质鲁棒性。
%表x 评估了 $\ours$ 在 CW 和 DI 攻击下的泛化性。结果显示，$\ours$ 在不同协议下均表现出卓越的防御力：在 CW 攻击下，其平均精度达到 54.0%，在 Pets 和 Flower102 等数据集上显著优于基线；在更具挑战性的 DI 攻击下，$\ours$ 以 46.2% 的精度保持领先，高出 R-TPT 2.0%。这表明 $\ours$ 并未过拟合于特定扰动模式，而是通过有效的测试时自适应机制，增强了预训练视觉-语言模型对抗多样化威胁的本质鲁棒性.
Table~\ref{tab:res_attack_type} further assesses the generalizability of $\ours$ against CW~\cite{atk-CW} and DI~\cite{atk-DI} attacks. $\ours$ consistently demonstrates superior defense across protocols, achieving a leading average accuracy of 54.0\% under CW attacks. Under the more challenging DI attack, $\ours$ maintains its advantage with 46.2\% accuracy, surpassing R-TPT by 2.0\%. These results suggest that $\ours$ enhances the intrinsic robustness of VLMs against diverse adversarial threats via its effective test-time adaptation mechanism.

\input{Figures/ablation_prompt_view}
\subsection{Ablation Study}
\noindent{\textbf{Number of learnable prompts and augmented views.}}
%learnbale prompt的数量，画柱状图（比如4,8, 16）一个数据集(同时clean和robust，就会有6个柱子)
%attack 图像增强的数量，画折线图(16,32,64,128, clean和robust, 两条线)
%上面两张图，单栏并排放
%我们在 DTD 数据集上针对可学习提示词数量及增强视图数量进行了敏感性分析，结果如图 2 所示。图 2(a) 展示了可学习提示词规模对模型性能的影响。实验结果表明，随着提示词数量从 4 增加至 32，模型在两种环境下的准确率均稳步增长。图 2(b) 评估了推理阶段增强视图规模的敏感性。观测结果显示，对抗精度随视图数量的增加而提升，并在达到 96 时达到峰值。然而，当视图数量进一步增至 128 时，性能出现了轻微回落，且这一趋势在无干扰环境下同样得到了印证。因此，选择适中的增强视图规模，能够实现在效率与精度之间的最佳权衡。
As illustrated in Figure~\ref{fig:ablation_prompt_view}, we conduct sensitivity analyses on the DTD dataset regarding the number of learnable prompts and augmentation views. Figure~\ref{fig:prompt_num} shows that increasing the number of learnable prompts from 4 to 32 leads to steady performance gains across both settings. Figure~\ref{fig:aug_view} assesses the sensitivity of augmentation view scales during inference, where adversarial accuracy peaks at 96 views. Notably, scaling to 128 views results in a slight performance decline, a trend consistently observed in noise-free environments as well. Consequently, selecting a moderate number of augmentation views enables a superior trade-off between efficiency and accuracy.

\noindent{\textbf{Contribution coefficient of dynamic cache.}}
%融合系数alpha,两个数据集，参考Enhancing CLIP Robustness via Cross-Modality Alignment中的Figure 3
%为了探究我们的方法对于推理阶段融合系数alpha的敏感性，我们在DTD数据集以及Imagenet数据集进行了实验，结果如图3所示。实验结果表明，引入缓存信息能有效提升模型性能。在两种环境下，alpha 从 0增至0.1时 ，两个数据集的准确率均较大提升；随后随着alpha的增加，准确率呈现小幅上升，趋于稳定。以上结果说明，缓存中的视觉先验不仅在对抗防御中起到了关键的修正作用，在标准推理场景下也能与原始逻辑值形成有效的语义补充。
We investigate the sensitivity of $\ours$ to the integration coefficient $\alpha$ during inference on the DTD and ImageNet datasets, as shown in Figure~\ref{fig:ablation_cache_alpha}. Experimental results confirm that the cache mechanism effectively boosts performance. On both datasets, increasing $\alpha$ from 0 to 0.1 yields a significant accuracy gain, followed by a slight upward trend that gradually stabilizes as $\alpha$ further scales. These results indicate that the visual priors stored in the cache not only play a crucial role in correcting predictions under adversarial settings, but also offer effective semantic complementarity during the inference stage. Ablation study of the dynamic cache is provided in Appendix~\ref{app_cache}.
\input{Figures/ablation_cache_alpha}

\noindent{\textbf{Different perturbation budgets.}}
%TAPT中的Figure5
%为了评估 RITA 在不同攻击强度下的稳健性，我们在 所有fine-grained 数据集 和 ImageNet 数据集上进行了消融实验，结果如 Figure 4 所示。随着干扰强度 $\epsilon$ 的增大（从 1 增加到 4），精度如预期发生下降，说明产生了更强的扰动。同时我们还发现增加 TTA 步数通常能带来额外的防御收益。
To assess the resilience of $\ours$ under various attack intensities, we conduct ablation studies on all fine-grained datasets and ImageNet dataset, with the results illustrated in Figure~\ref{fig:ablation_epsilon}. As the perturbation budget $\epsilon$ increases from 1 to 4, the accuracy declines as expected, indicating the generation of more potent perturbations. Simultaneously, we observe that increasing the number of TTA steps generally yields additional and consistent defensive gains across evaluations. 
\input{Figures/ablation_epsilon}

\noindent{\textbf{Number of TTA steps.}}
%TAPT中的Figure4,只要(a)average和(b)ImageNet，其中Average不包含ImageNet
%我们给出了三种骨架下，所有细粒度数据集的平均鲁棒精度（见图5（a））以及Imagenet数据集的平均鲁棒精度（见图5（b）），TTA step=0代表没有RITA的基线，即原始CLIP的表现。实验结果表明，各模型在第 2 步时均达到性能峰值，而增加到 4 步时精度略有回落。这一观察结果证明了我们选择少量迭代次数（TTA step=1）的合理性，既确保了卓越的鲁棒性，又保证了实时推理的高计算效率。
We report the average robust accuracy across three different backbones on all fine-grained datasets as shown in Figure~\ref{fig:TTAstep_FG} and the ImageNet dataset as illustrated in Figure~\ref{fig:TTAstep_Imagenet}. The experimental results indicate that all models reach their performance peak at step 2, while the accuracy slightly declines when the steps are increased to 4 across all evaluated datasets. This observation justifies our choice of a small number of iterations, \eg, TTA step = 1, which simultaneously ensures superior robustness and high computational efficiency for real-time inference.
\input{Figures/ablation_TTAstep}

\noindent{\textbf{Inference efficiency analysis.}}
%表5展示了不同方法在细粒度数据集上的运行时间、干净环境精度（Clean）以及鲁棒精度（Robust）的平均对比结果。实验结果表明，RITA 在保持极高竞争力的推理效率的同时，实现了最优的分类性能。与推理速度最快的 MTA相比，RITA 在鲁棒精度上领先了 11.8%。这证明了 RITA在计算成本与模型鲁棒性之间取得了卓越的平衡，能够以极小的额外时间成本为代价，换取更稳健、更精确的推理结果。
Table~\ref{tab:ablation_running_time} presents the average comparison of running time, clean accuracy, and robust accuracy across various methods on fine-grained datasets. The empirical results indicate that $\ours$ achieves superior classification performance while maintaining highly competitive inference efficiency compared with existing adaptation methods. Compared to MTA, which exhibits the fastest inference speed, $\ours$ maintains a substantial lead of 11.8\% in robust accuracy. These observations demonstrate that $\ours$ strikes an excellent balance between computational cost and model robustness, delivering more resilient and precise inference results with minimal additional time latency.

\input{Tables/ablation_running-time}

\noindent{\textbf{Semantic alignment analysis.}}
%如图 X 所示，我们通过 KDE 曲线直观地展示了引入 Cache 机制前后语义对齐质量的演变。对于每个类别，我们计算得到文本原型上的类条件分配分布 $\tilde{\mathbb{P}}$，并计算其与理想单热点（One-hot）目标分布 $\tilde{\mathbb{P}}_{\mathrm{gt}}$之间的 Kullback-Leibler (KL) 散度。在 Caltech101 和 Flower102 两个数据集上，增强特征对应的 KL 散度曲线相较于原始特征均表现出显著的左移趋势，且分布更加向低值区域集中。这种现象定量地证明了：通过多视图增强与 Cache 机制的结合，模型能够显著修正对抗环境下的语义偏差，从而建立一种更具确定性且更稳定的视觉-文本关联。
As shown in Figure~\ref{fig:ablation_alignment}, we visualize semantic alignment quality via KDE curves of the Kullback-Leibler (KL) divergence. Specifically, we measure the divergence between the class-conditional distributions over text prototypes and the ideal one-hot targets. On both DTD and Caltech101 datasets, the curves for augmented features exhibit a pronounced leftward shift and higher concentration in the low-value region compared to the original features. This demonstrates that our cache mechanism effectively rectifies adversarial semantic biases, establishing more deterministic vision-text associations. Extended analysis is provided in Appendix~\ref{app_alignment}.
\input{Figures/ablation_alignment}

%相似度分布或者距离分布，参考参考Enhancing CLIP Robustness via Cross-Modality Alignment中的Figure 3中的Figure 4

%% file: Tables/res_vitb32-16.tex
\begin{table*}[!t]
\centering
\tabstyle{2.5pt}
\caption{Results (\%) of adaptation methods on fine-grained classification datasets with $\epsilon$ set to 1.0. \bestclean{Bold} and \underline{underlined} entries indicate the best and second-best results, respectively. 
Acc. denotes accuracy on clean data, and Rob. denotes accuracy under adversarial perturbations.
}
\begin{tabular}{c| l|cc|cc|cc|cc|cc|cc|cc|cc|cc}
\toprule
\multirow{2}{*}{} & 
\multirow{2}{*}{Method} & 
\multicolumn{2}{c|}{Caltech101} & 
\multicolumn{2}{c|}{Pets} & 
\multicolumn{2}{c|}{Cars} & 
\multicolumn{2}{c|}{Flower102} & 
\multicolumn{2}{c|}{Aircraft} & 
\multicolumn{2}{c|}{DTD} & 
\multicolumn{2}{c|}{EuroSAT} & 
\multicolumn{2}{c|}{UCF101} & 
\multicolumn{2}{c}{Avg.} \\
 &  & 
Acc. & Rob. & 
Acc. & Rob. & 
Acc. & Rob. & 
Acc. & Rob. & 
Acc. & Rob. & 
Acc. & Rob. & 
Acc. & Rob. & 
Acc. & Rob. & 
Acc. & Rob. \\
\midrule
\multirow{7}{*}{\rotatebox{90}{ViT-B/32}} 
& CLIP 
& 90.9 & 25.6 & 83.0 & 2.1 & 49.7 & 0.0 & 65.8 & 2.2 & 18.3 & 0.0 & 40.8 & 5.4 & 18.6 & 0.1 & 62.1 & 2.4 & 53.6 & 4.7  \\

& Ensemble 
& 91.6 & 85.4 & 85.0 & \underline{73.4} & 57.8 & 39.6 & \underline{67.4} & \underline{57.5} & 20.1 & \underline{14.4} & \underline{46.1} & \underline{39.3} & 32.5 & \underline{23.6} & 61.6 & 53.5 & \underline{57.8} & \underline{48.3}  \\

& TPT
& 91.4 & 77.6 & 84.1 & 58.6 & 62.9 & 30.8 & 63.8 & 43.6 & 19.0 & 10.1 & 42.2 & 28.6 & \bestclean{35.1} & 15.1 & 62.3 & 38.1 & 57.6 & 37.8  \\

& C-TPT
& \underline{91.8} & 75.9 & 84.9 & 52.2 & 60.8 & 27.1 & 65.9 & 42.1 & 17.7 & 8.7 & 44.3 & 27.1 & \underline{34.7} & 9.0 & 62.6 & 35.3 & \underline{57.8} & 34.6  \\

& MTA
& \underline{91.8} & 80.8 & \underline{85.8} & 62.6 & \bestclean{64.1} & 34.5 & 64.8 & 44.8 & \bestclean{20.4} & 11.1 & 44.0 & 29.3 & 34.5 & 7.8 & \bestclean{63.6} & 40.1 & \bestclean{58.6} & 38.8  \\

& R-TPT
& 90.6 & \underline{86.2} & 84.5 & 73.1 & \underline{63.1} & \bestclean{44.6} & 62.6 & 53.1 & 19.1 & 12.9 & 42.1 & 36.7 & 32.0 & 22.4 & \underline{62.8} & \underline{54.2} & 57.1 & 47.9  \\

& \cellcolor{rowgreen}\ours
& \cellcolor{rowgreen}\bestclean{92.3} 
& \cellcolor{rowgreen}\bestclean{86.5} 
& \cellcolor{rowgreen}\bestclean{85.9} 
& \cellcolor{rowgreen}\bestclean{74.5} 
& \cellcolor{rowgreen}59.6 
& \cellcolor{rowgreen}\underline{42.9} 
& \cellcolor{rowgreen}\bestclean{68.7} 
& \cellcolor{rowgreen}\bestclean{58.7} 
& \cellcolor{rowgreen}\underline{20.2} 
& \cellcolor{rowgreen}\bestclean{15.2} 
& \cellcolor{rowgreen}\bestclean{46.2} 
& \cellcolor{rowgreen}\bestclean{40.1} 
& \cellcolor{rowgreen}33.4 
& \cellcolor{rowgreen}\bestclean{24.8} 
& \cellcolor{rowgreen}\underline{62.8} 
& \cellcolor{rowgreen}\bestclean{55.0} 
& \cellcolor{rowgreen}\bestclean{58.6} 
& \cellcolor{rowgreen}\bestclean{49.7}  \\ \midrule

\multirow{7}{*}{\rotatebox{90}{ViT-B/16}} 
& CLIP
& 85.9 & 10.8 & 83.5 & 0.5 & 55.7 & 0.0 & 61.7 & 0.1 & 15.7 & 0.0 & 40.4 & 2.4 & 23.7 & 0.0 & 58.9 & 0.5 & 53.2 & 1.8  \\

& Ensemble 
& 92.1 & 87.4 & \underline{88.7} & \underline{77.2} & 63.2 & 46.7 & \underline{70.8} & \underline{59.9} & \underline{25.9} & \underline{17.9} & \underline{50.9} & \underline{43.2} & 32.9 & 26.7 & 64.6 & 54.3 & 61.1 & \underline{51.6}  \\

& TPT
& \underline{94.1} & 79.6 & 87.4 & 62.8 & 66.5 & 35.5 & 66.1 & 48.3 & 23.4 & 12.3 & 45.9 & 29.1 & \bestclean{42.6} & 7.4 & \bestclean{67.9} & 39.7 & 61.7 & 39.3  \\

& C-TPT
& 93.9 & 76.5 & 88.2 & 55.8 & 65.8 & 30.5 & 69.6 & 45.5 & 23.9 & 9.8 & 45.9 & 26.6 & 42.3 & 7.1 & 65.6 & 34.7 & \underline{61.9} & 35.8  \\

& MTA
& \bestclean{94.3} & 81.9 & 88.0 & 64.5 & \bestclean{67.7} & 38.2 & 65.0 & 46.9 & 24.0 & 12.6 & 46.5 & 28.7 & \underline{42.5} & 13.7 & \underline{67.5} & 40.8 & \underline{61.9} & 40.9  \\

& R-TPT 
& 93.7 & \underline{87.8} & 87.2 & 74.7 & \underline{67.0} & \underline{46.9} & 68.7 & 55.7 & 23.9 & 17.3 & 46.4 & 39.7 & 34.7 & \underline{26.8} & 67.2 & \underline{55.4} & 61.1 & 50.5  \\

& \cellcolor{rowgreen}\ours
& \cellcolor{rowgreen}93.8 & \cellcolor{rowgreen}\bestclean{88.5} & \cellcolor{rowgreen}\bestclean{89.8} & \cellcolor{rowgreen}\bestclean{77.3} & \cellcolor{rowgreen}64.2 & \cellcolor{rowgreen}\bestclean{47.1} & \cellcolor{rowgreen}\bestclean{71.6} & \cellcolor{rowgreen}\bestclean{61.3} & \cellcolor{rowgreen}\bestclean{26.2} & \cellcolor{rowgreen}\bestclean{19.2} & \cellcolor{rowgreen}\bestclean{51.5} & \cellcolor{rowgreen}\bestclean{44.7} & \cellcolor{rowgreen}33.4 & \cellcolor{rowgreen}\bestclean{27.6} & \cellcolor{rowgreen}65.5 & \cellcolor{rowgreen}\bestclean{55.8} & \cellcolor{rowgreen}\bestclean{62.0} & \cellcolor{rowgreen}\bestclean{52.7}  \\

\bottomrule
\end{tabular}
\label{tab:res_vitb32-16}
\end{table*}

%% file: Tables/ablation_finetuning.tex
\begin{table*}[t]
\caption{Classification accuracy (\%) on 8 datasets using different adversarially finetuned CLIP models. 
% \bestclean{Bold} and \underline{underlined} denote best and second-best.
}
\label{tab:fine-t_9datasets}
\centering
\tabstyle{3pt}
\begin{tabular}{c|cc|cc|cc|cc|cc|cc|cc|cc|cc}
\toprule
\multirow{2}{*}{\rotatebox{0}{Method}} & \multicolumn{2}{c|}{\rotatebox{0}{Caltech101}} & \multicolumn{2}{c|}{\rotatebox{0}{Pets}} & \multicolumn{2}{c|}{\rotatebox{0}{Cars}} & \multicolumn{2}{c|}{\rotatebox{0}{Flower102}} & \multicolumn{2}{c|}{\rotatebox{0}{Aircraft}} & \multicolumn{2}{c|}{\rotatebox{0}{DTD}} & \multicolumn{2}{c|}{\rotatebox{0}{EuroSAT}} & \multicolumn{2}{c|}{\rotatebox{0}{UCF101}}  & \multicolumn{2}{c}{\rotatebox{0}{Avg.}}  \\ %\cmidrule{2-19}
& Acc. & Rob. 
& Acc. & Rob. 
& Acc. & Rob.
& Acc. & Rob.
& Acc. & Rob.
& Acc. & Rob.
& Acc. & Rob.
& Acc. & Rob.
& Acc. & Rob. \\
\midrule
TeCoA & 77.6 & 64.3 & 59.8 & 39.9 & 20.6 & 9.3 & 37.1 & 23.9 & 5.7 & 2.8 & 24.1 & 14.4 & \bestclean{15.9} & \bestclean{12.4} & 40.5 & 23.4 & 35.2 & 23.8  \\
+ Ensemble & 76.0 & 68.3 & 61.3 & \underline{54.8} & 19.4 & 14.5 & \underline{38.4} & \underline{33.2} & 6.9 & \underline{4.9} & \underline{27.5} & \underline{25.2} & 11.5 & 12.0 & 39.4 & 34.5 & 35.1 & 30.9  \\
+ MTA & \bestclean{79.5} & 56.5 & \underline{61.6} & 28.8 & 20.9 & 10.4 & 37.2 & 24.0 & 6.6 & 3.8 & 25.8 & 19.7 & \underline{12.2} & 11.2 & \bestclean{42.0} & 17.3 & \underline{35.7} & 21.4 \\
+ R-TPT & 77.1 & \underline{70.3} & 60.9 & 54.0 & \bestclean{23.1} & \bestclean{17.8} & 35.4 & 30.8 & \underline{7.0} & 4.7 & 26.7 & 24.7 & 12.0 & 11.8 & 40.1 & \underline{35.0} & 35.3 & \underline{31.1}  \\
+ \cellcolor{rowgreen}{\ours}  & \cellcolor{rowgreen}\underline{78.1} & \cellcolor{rowgreen}\bestclean{70.4} & \cellcolor{rowgreen}\bestclean{62.3} & \cellcolor{rowgreen}\bestclean{54.9} & \cellcolor{rowgreen}\underline{21.1} & \cellcolor{rowgreen}\underline{15.7} & \cellcolor{rowgreen}\bestclean{39.2} & \cellcolor{rowgreen}\bestclean{34.0} & \cellcolor{rowgreen}\bestclean{7.1} & \cellcolor{rowgreen}\bestclean{5.5} & \cellcolor{rowgreen}\bestclean{27.8} & \cellcolor{rowgreen}\bestclean{25.5} & \cellcolor{rowgreen}\underline{12.2} & \cellcolor{rowgreen}\underline{12.1} & \cellcolor{rowgreen}\underline{40.9} & \cellcolor{rowgreen}\bestclean{35.8} & \cellcolor{rowgreen}\bestclean{36.1} & \cellcolor{rowgreen}\bestclean{31.7}  \\ \midrule
PMG & \bestclean{82.3} & 70.8 & 61.8 & 41.4 & \bestclean{24.7} & 12.5 & 36.2 & 25.3 & 5.2 & 2.9 & 22.8 & 16.0 & \bestclean{17.1} & 12.8 & 42.6 & 27.8 & \underline{36.5} & 26.1  \\
+ Ensemble & 78.9 & 71.7 & 60.7 & 54.1 & 16.6 & 11.6 & \underline{37.8} & \underline{32.7} & \underline{7.2} & \underline{5.1} & \underline{26.6} & \underline{25.2} & 14.0 & \underline{13.9} & 42.0 & 37.7 & 35.5 & \underline{31.5}  \\
+ MTA & 79.5 &  65.4 & 61.8 & 31.3 & 17.9 & 12.9 & 36.7 & 21.6 & 5.8 & 3.3 & 22.9 & 18.7 & 13.8 & 12.5 & \underline{43.1} & 22.6 & 35.1 & 23.5  \\
+ R-TPT & 79.3 & \underline{73.2} & \underline{62.0} & \underline{55.1} & 18.3 & \bestclean{14.9} & 35.3 & 30.4 & 5.4 & 4.1 & 25.4 & 23.0 & 13.2 & 12.9 & 42.3 & \underline{38.2} & 35.2 & 31.4  \\
+ \cellcolor{rowgreen}{\ours} & \cellcolor{rowgreen}\underline{79.9} & \cellcolor{rowgreen}\bestclean{73.7} & \cellcolor{rowgreen}\bestclean{62.5} & \cellcolor{rowgreen}\bestclean{55.4} & \cellcolor{rowgreen}\underline{18.9} & \cellcolor{rowgreen}\underline{13.9} & \cellcolor{rowgreen}\bestclean{38.7} & \cellcolor{rowgreen}\bestclean{33.9} & \cellcolor{rowgreen}\bestclean{7.5} & \cellcolor{rowgreen}\bestclean{5.4} & \cellcolor{rowgreen}\bestclean{27.4} & \cellcolor{rowgreen}\bestclean{25.4} & \cellcolor{rowgreen}\underline{14.4} & \cellcolor{rowgreen}\bestclean{14.1} & \cellcolor{rowgreen}\bestclean{43.5} & \cellcolor{rowgreen}\bestclean{38.6} & \cellcolor{rowgreen}\bestclean{36.6} & \cellcolor{rowgreen}\bestclean{32.6}  \\ \midrule
FARE & 86.6 & 62.9 & 77.7 & 38.1 & 40.4 & 9.7 & 48.7 & 22.5 & 10.2 & 2.3 & 32.4 & 18.1 & \bestclean{22.4} & 11.0 & 52.9 & 22.2 & \underline{46.4} & 23.3  \\
+ Ensemble & 86.1 & 80.2 & 77.9 & \underline{70.1} & 38.8 & 29.5 & 48.9 & \underline{42.3} & 10.5 & \underline{7.8} & \underline{36.9} & \underline{32.4} & 13.5 & 11.6 & 52.8 & 45.6 & 45.6 & 39.9  \\
+ MTA & \bestclean{87.7} & 70.0 & \underline{78.4} & 45.0 & 40.6 & 24.8 & \underline{49.2} & 30.5 & \underline{11.0} & 6.6 & 32.8 & 25.3 & 13.5 & 11.8 & \underline{53.9} & 29.8 & 45.8 & 30.4  \\
+ R-TPT & 86.5 & \underline{81.1} & 77.4 & \underline{70.1} & \bestclean{43.0} & \bestclean{33.0} & 46.2 & 40.2 & 10.0 & 7.4 & 33.6 & 30.0 & 12.9 & \underline{12.1} & 53.8 & \underline{46.8} & 45.4 & \underline{40.0}  \\
+ \cellcolor{rowgreen}{\ours} & \cellcolor{rowgreen}\underline{86.8} & \cellcolor{rowgreen}\bestclean{81.9} & \cellcolor{rowgreen}\bestclean{78.7} & \cellcolor{rowgreen}\bestclean{70.5} & \cellcolor{rowgreen}\underline{41.2} & \cellcolor{rowgreen}\underline{31.2} & \cellcolor{rowgreen}\bestclean{50.1} & \cellcolor{rowgreen}\bestclean{43.1} & \cellcolor{rowgreen}\bestclean{11.8} & \cellcolor{rowgreen}\bestclean{8.6} & \cellcolor{rowgreen}\bestclean{37.5} & \cellcolor{rowgreen}\bestclean{32.9} & \cellcolor{rowgreen}\underline{14.2} & \cellcolor{rowgreen}\bestclean{12.7} & \cellcolor{rowgreen}\bestclean{54.3} & \cellcolor{rowgreen}\bestclean{47.0} & \cellcolor{rowgreen}\bestclean{46.8} & \cellcolor{rowgreen}\bestclean{41.0}   \\
\bottomrule
\end{tabular}
\label{tab:ablation_finetuning}
\end{table*}

%% file: Tables/res_attack_type.tex
\begin{table*}[ht]
\centering
\tabstyle{3pt}
\caption{Results (\%) of adaptation methods on fine-grained classification datasets under different attacks using ViT-B/16 with $\epsilon$ = 1.0. 
% \bestclean{Bold} and \underline{underlined} denote best and second-best.
}
\begin{tabular}{c| l|cc|cc|cc|cc|cc|cc|cc|cc|cc}
\toprule
\multirow{2}{*}{} & 
\multirow{2}{*}{Method} & 
\multicolumn{2}{c|}{Caltech101} & 
\multicolumn{2}{c|}{Pets} & 
\multicolumn{2}{c|}{Cars} & 
\multicolumn{2}{c|}{Flower102} & 
\multicolumn{2}{c|}{Aircraft} & 
\multicolumn{2}{c|}{DTD} & 
\multicolumn{2}{c|}{EuroSAT} & 
\multicolumn{2}{c|}{UCF101} & 
\multicolumn{2}{c}{Avg.} \\
 &  & 
Acc. & Rob. & 
Acc. & Rob. & 
Acc. & Rob. & 
Acc. & Rob. & 
Acc. & Rob. & 
Acc. & Rob. & 
Acc. & Rob. & 
Acc. & Rob. & 
Acc. & Rob. \\
\midrule
\multirow{7}{*}{\rotatebox{90}{CW}} 
& CLIP
& 85.9 & 22.7 & 83.5 & 7.7 & 55.7 & 6.6 & 61.7 & 5.8 & 15.7 & 8.8 & 40.4 & 16.3 & 23.7 & 15.4 & 58.9 & 11.3 & 53.2 & 11.8  \\

& Ensemble 
& 92.1 & 86.7 & \underline{88.7} & \underline{78.6} & 63.2 & 50.5 & \underline{70.8} & \underline{61.6} & \underline{25.9} & \underline{22.2} & \underline{50.9} & \underline{45.2} & 32.9 & \underline{24.8} & 64.6 & 55.3 & 61.1 & \underline{53.1}  \\

& TPT
& \underline{94.1} & 77.2 & 87.4 & 66.9 & 66.5 & 43.2 & 66.1 & 49.9 & 23.4 & 16.1 & 45.9 & 30.7 & \bestclean{42.6} & 13.5 & \bestclean{67.9} & 44.1 & 61.7 & 42.7  \\

& C-TPT
& 93.9 & 76.0 & 88.2 & 62.1 & 65.8 & 39.6 & 69.6 & 47.5 & 23.9 & 15.1 & 45.9 & 30.2 & 42.3 & 11.5 & 65.6 & 40.3 & \underline{61.9} & 40.2  \\

& MTA 
& \bestclean{94.3} & 79.2 & 88.0 & 67.8 & \bestclean{67.7} & 43.4 & 65.0 & 48.4 & 24.0 & 16.1 & 46.5 & 30.6 & \underline{42.5} & 19.3 & \underline{67.5} & 44.6 & \underline{61.9} & 43.6  \\

& R-TPT 
& 93.7 & \bestclean{88.1} & 87.2 & 74.4 & \underline{67.0} & \underline{50.7} & 68.7 & 55.7 & 23.9 & 20.1 & 46.4 & 39.6 & 34.7 & \underline{24.8} & 67.2 & \underline{56.2} & 61.1 & 51.2  \\

& \cellcolor{rowgreen}\ours
& \cellcolor{rowgreen}93.8 
& \cellcolor{rowgreen}\underline{87.8} 
& \cellcolor{rowgreen}\bestclean{89.8} 
& \cellcolor{rowgreen}\bestclean{78.7} 
& \cellcolor{rowgreen}64.2 
& \cellcolor{rowgreen}\bestclean{52.6} 
& \cellcolor{rowgreen}\bestclean{71.6} 
& \cellcolor{rowgreen}\bestclean{62.0} 
& \cellcolor{rowgreen}\bestclean{26.2} 
& \cellcolor{rowgreen}\bestclean{22.9} 
& \cellcolor{rowgreen}\bestclean{51.5} 
& \cellcolor{rowgreen}\bestclean{46.2} 
& \cellcolor{rowgreen}33.4 
& \cellcolor{rowgreen}\bestclean{25.2} 
& \cellcolor{rowgreen}65.5 
& \cellcolor{rowgreen}\bestclean{56.8} 
& \cellcolor{rowgreen}\bestclean{62.0}
& \cellcolor{rowgreen}\bestclean{54.0} \\ \midrule

\multirow{7}{*}{\rotatebox{90}{ DI}} 
& CLIP 
& 85.9 & 23.0 & 83.5 & 4.4 & 55.7 & 0.6 & 61.7 & 1.9 & 15.7 & 0.0 & 40.4 & 6.1 & 23.7 & 0.0 & 58.9 & 3.2 & 53.2 & 4.9  \\

& Ensemble 
& 92.1 & 84.3 & \underline{88.7} & \underline{69.1} & 63.2 & 39.1 & \underline{70.8} & \underline{52.7} & \underline{25.9} & \underline{15.3} & \underline{50.9} & \underline{39.7} & 32.9 & 17.6 & 64.6 & \underline{47.6} & 61.1 & \underline{45.7}  \\

& TPT
& \underline{94.1} & 80.7 & 87.4 & 65.2 & 66.5 & 38.3 & 66.1 & 49.7 & 23.4 & 13.5 & 45.9 & 30.3 & \bestclean{42.6} & 7.4 & \bestclean{67.9} & 40.5 & 61.7 & 40.7  \\

& C-TPT 
& 93.9 & 79.5 & 88.2 & 59.5 & 65.8 & 34.2 & 69.6 & 47.3 & 23.9 & 11.6 & 45.9 & 29.1 & 42.3 & 7.4 & 65.6 & 37.0 & \underline{61.9} & 38.2  \\

& MTA
& \bestclean{94.3} & 82.6 & 88.0 & 65.6 & \bestclean{67.7} & \underline{39.5} & 65.0 & 48.0 & 24.0 & 13.5 & 46.5 & 30.9 & \underline{42.5} & 14.7 & \underline{67.5} & 41.9 & \underline{61.9} & 42.1  \\

& R-TPT 
& 93.7 & \bestclean{84.9} & 87.2 & 66.7 & \underline{67.0} & 39.1 & 68.7 & 48.1 & 23.9 & 14.1 & 46.4 & 35.6 & 34.7 & \underline{18.0} & 67.2 & 47.1 & 61.1 & 44.2  \\

& \cellcolor{rowgreen}\ours
& \cellcolor{rowgreen}93.8 
& \cellcolor{rowgreen}\underline{84.7} 
& \cellcolor{rowgreen}\bestclean{89.8} 
& \cellcolor{rowgreen}\bestclean{69.4} 
& \cellcolor{rowgreen}64.2 
& \cellcolor{rowgreen}\bestclean{39.9} 
& \cellcolor{rowgreen}\bestclean{71.6} 
& \cellcolor{rowgreen}\bestclean{53.4} 
& \cellcolor{rowgreen}\bestclean{26.2} 
& \cellcolor{rowgreen}\bestclean{15.9} 
& \cellcolor{rowgreen}\bestclean{51.5} 
& \cellcolor{rowgreen}\bestclean{39.8} 
& \cellcolor{rowgreen}33.4 
&\cellcolor{rowgreen}\bestclean{18.4} 
& \cellcolor{rowgreen}65.5 
& \cellcolor{rowgreen}\bestclean{47.7} 
& \cellcolor{rowgreen}\bestclean{62.0} 
& \cellcolor{rowgreen}\bestclean{46.2} \\

\bottomrule
\end{tabular}
\label{tab:res_attack_type}
\end{table*}

%% file: Figures/ablation_prompt_view.tex
\begin{figure}[!t]
    \centering
    % 第一张子图
    \begin{subfigure}{0.48\columnwidth}
        \centering
        \includegraphics[width=\linewidth]{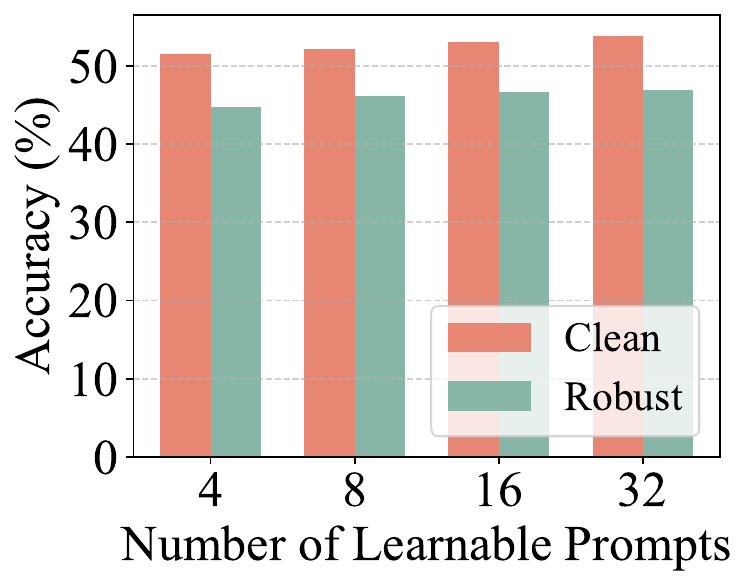}
        \caption{Prompt Number}
        \label{fig:prompt_num}
    \end{subfigure}
    \hfill % 在两图之间填充空白
    % 第二张子图
    \begin{subfigure}{0.48\columnwidth}
        \centering
        \includegraphics[width=\linewidth]{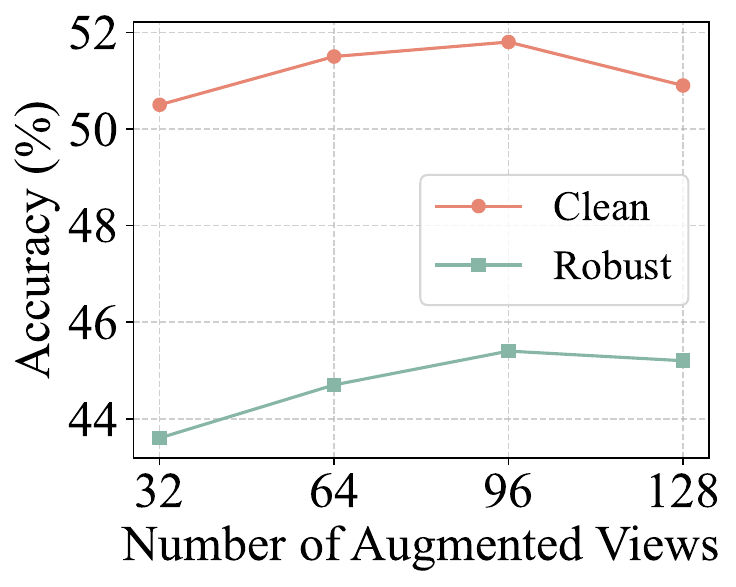}
        \caption{Augmentation View}
        \label{fig:aug_view}
    \end{subfigure}
    
    \caption{Ablation studies of key hyperparameters in DTD dataset using ViT-B/16 with $\epsilon$ = 1.0. (a) Performance variation with respect to the number of learnable prompts. (b) Impact of the number of augmentation views during inference. } % 总标题
    \label{fig:ablation_prompt_view}
\end{figure}

%% file: Figures/ablation_cache_alpha.tex
\begin{figure}[!t]
    \centering
    % 第一张子图
    \begin{subfigure}{0.48\columnwidth}
        \centering
        \includegraphics[width=\linewidth]{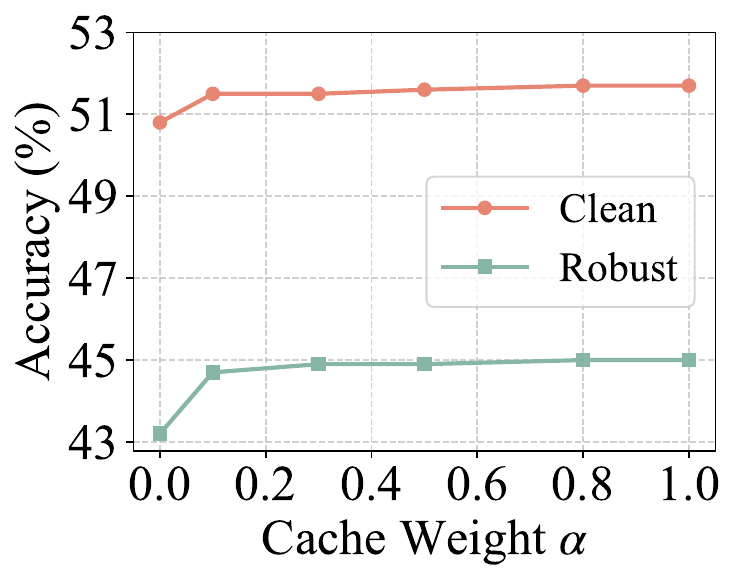}
        \caption{DTD}
        \label{fig:cache_alpha_dtd}
    \end{subfigure}
    \hfill % 在两图之间填充空白
    % 第二张子图
    \begin{subfigure}{0.48\columnwidth}
        \centering
        \includegraphics[width=\linewidth]{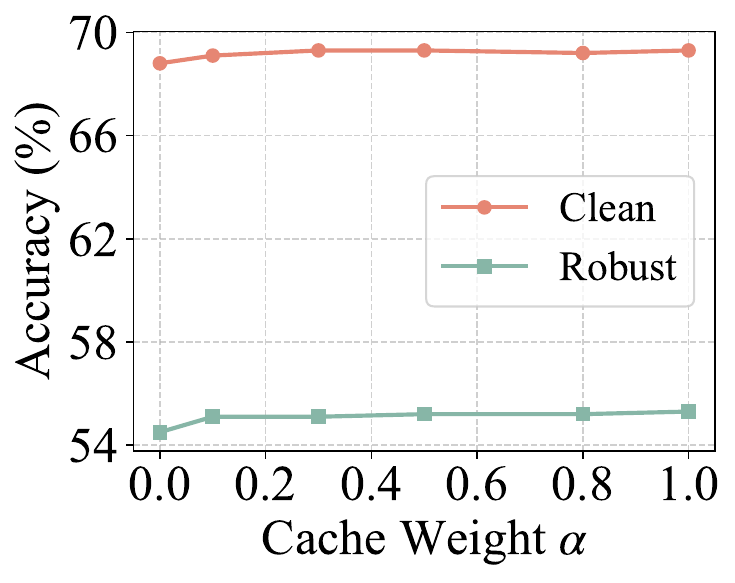}
        \caption{ImageNet}
        \label{fig:cache_alpha_imagenet}
    \end{subfigure}
    \caption{
    Sensitivity analysis of the cache integration coefficient $\alpha$. Classification accuracy (\%) on (a) DTD and (b) ImageNet is evaluated across $\alpha$ values using ViT-B/16 with $\epsilon$ = 1.0.
    } % 总标题
    \label{fig:ablation_cache_alpha}
\end{figure}

%% file: Figures/ablation_epsilon.tex
\begin{figure}[!t]
    \centering
    % 第一张子图
    \begin{subfigure}{0.48\columnwidth}
        \centering
        \includegraphics[width=\linewidth]{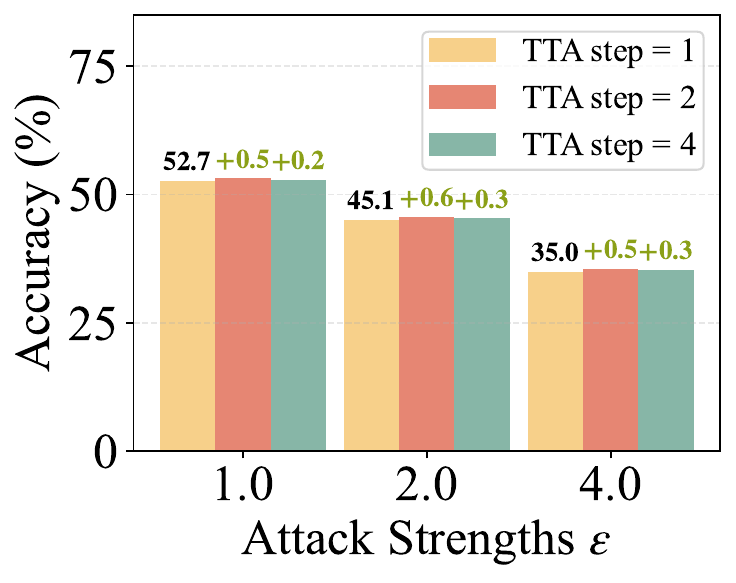}
        \caption{Fine-grained datasets}
        \label{fig:epsilon_FG}
    \end{subfigure}
    \hfill % 在两图之间填充空白
    % 第二张子图
    \begin{subfigure}{0.48\columnwidth}
        \centering
        \includegraphics[width=\linewidth]{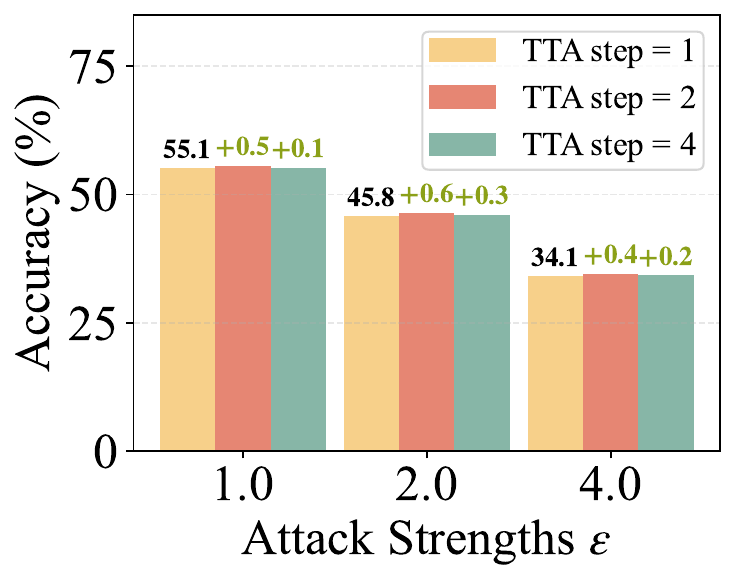}
        \caption{ImageNet}
        \label{fig:epsilon_imagenet}
    \end{subfigure}
    
    \caption{Adversarial robustness (\%) under varying perturbation budgets and TTA steps on (a) fine-grained datasets and (b) ImageNet. Robust accuracy is evaluated using ViT-B/16 under $\epsilon \in \{1.0, 2.0, 4.0\}$ and TTA steps $\in \{1, 2, 4\}$. \textcolor[HTML]{8BA018}{+x.x} indicates the increment relative to the case where TTA step = 1.} % 总标题
    \label{fig:ablation_epsilon}
\end{figure}

%% file: Figures/ablation_TTAstep.tex
\begin{figure}[!t]
    \centering
    % 第一张子图
    \begin{subfigure}{0.48\columnwidth}
        \centering
        \includegraphics[width=\linewidth]{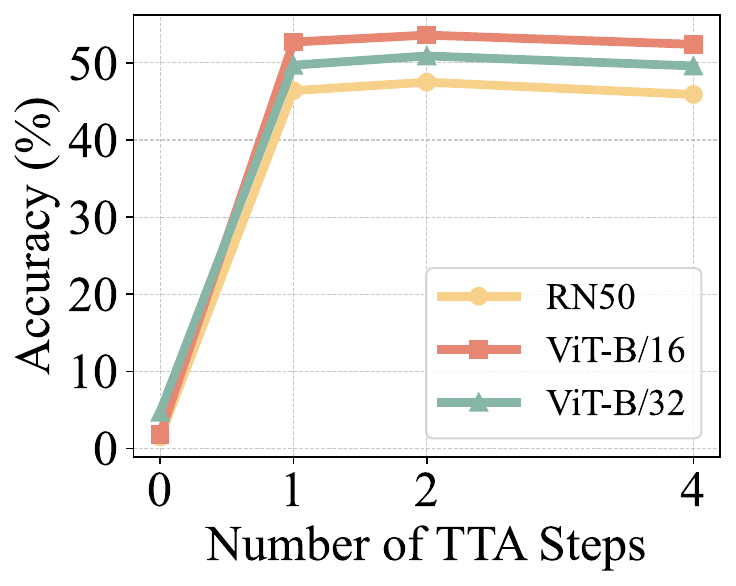}
        \caption{Fine-grained datasets}
        \label{fig:TTAstep_FG}
    \end{subfigure}
    \hfill % 在两图之间填充空白
    % 第二张子图
    \begin{subfigure}{0.48\columnwidth}
        \centering
        \includegraphics[width=\linewidth]{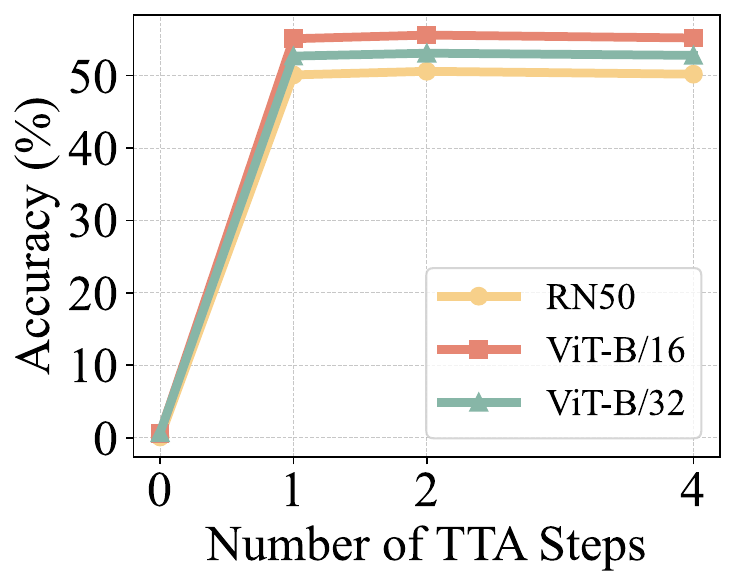}
        \caption{ImageNet}
        \label{fig:TTAstep_Imagenet}
    \end{subfigure}
    
    \caption{Evolution of adversarial robustness with respect to TTA steps across model architectures, where (a) presents the average robust accuracy over fine-grained datasets and (b) displays results on ImageNet. TTA step = 0 denotes the CLIP baseline.} % 总标题
    \label{fig:ablation_TTAstep}
    \vspace{-5pt}
\end{figure}

%% file: Tables/ablation_running-time.tex
\begin{table}[!t]
    \centering
\caption{Comparison of running time on fine-grained datasets using ViT-B/16 with $\epsilon$ = 1.0.}
% \vspace{-0.1cm}
\label{tab:time}
\tabstyle{12pt}
\centering
\begin{tabular}{lccc}
\toprule
\multirow{2}{*}{Model} & \multirow{2}{*}{Running Time} & \multicolumn{2}{c}{Accuracy} \\
\cmidrule(lr){3-4}
& & Clean & Robust \\
\midrule
TPT  & 1.52s/image  & 61.7 & 39.3 \\
C-TPT  & 1.64s/image  & 61.9 & 35.8 \\
MTA  & 1.20s/image  & 61.9 & 40.9 \\
R-TPT  & 1.70s/image  & 61.1 & 50.5 \\
{\ours} & 1.76s/image & \bestclean{62.0} & \bestclean{52.7} \\
\bottomrule
\end{tabular}
\label{tab:ablation_running_time}
\end{table}

% % \setlength{\tabcolsep}{4.0pt}
%     \begin{table}[!t]
%         \centering
%         \caption{ Clean and adversarial accuracies (\%) on fine-grained datasets and ImageNet dataset (CLIP-ResNet50).}
%         % \vspace{-3mm}
%         \centering
%         % \resizebox{\linewidth}{!}
%         % {
%             \begin{tabular}{cc|cccc}
%             \toprule
%             \multirow{2}{*}{$p(k \mid \hat{x}_t)$} & \multirow{2}{*}{$p_\mathrm{cache}(k \mid \hat{x}_t)$} & \multicolumn{2}{c}{Fine-grained}  & \multicolumn{2}{c}{ImageNet} \\
%              &   & Acc. & Rob. & Acc. & Rob. \\
%             \midrule
%             \CheckmarkBold & \XSolidBrush  & 50.6 & 40.4 & 42.7 & 29.5 \\
%             \XSolidBrush   & \CheckmarkBold & 53.3 & 44.1 & 46.7 & 34.2 \\
%             \CheckmarkBold & \CheckmarkBold  & 53.7 & \best{45.2} & 47.1 & \best{35.4} \\
%             \bottomrule
%             \end{tabular}
%         % }
%     \label{tab:ablation}
%     \end{table}

% p(k \mid \hat{x}_t) \leftarrow p(k \mid \hat{x}_t) + p_\mathrm{cache}(k \mid \hat{x}_t).

%% file: Figures/ablation_alignment.tex
\begin{figure}[!t]
    \centering
    % 第一张子图
    \begin{subfigure}{0.48\columnwidth}
        \centering
        \includegraphics[width=\linewidth]{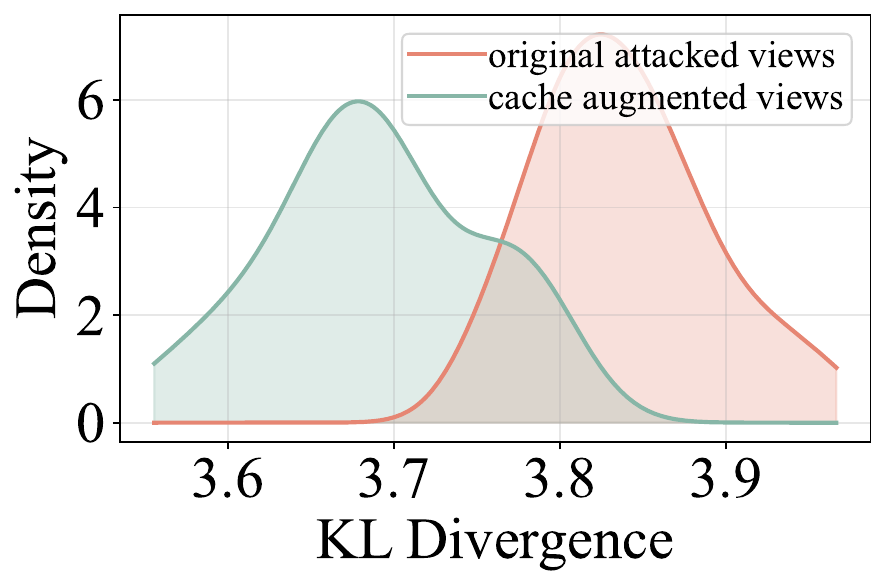}
        \caption{DTD}
        \label{fig:kl_density_DTD}
    \end{subfigure}
    \hfill % 在两图之间填充空白
    % 第二张子图
    \begin{subfigure}{0.48\columnwidth}
        \centering
        \includegraphics[width=\linewidth]{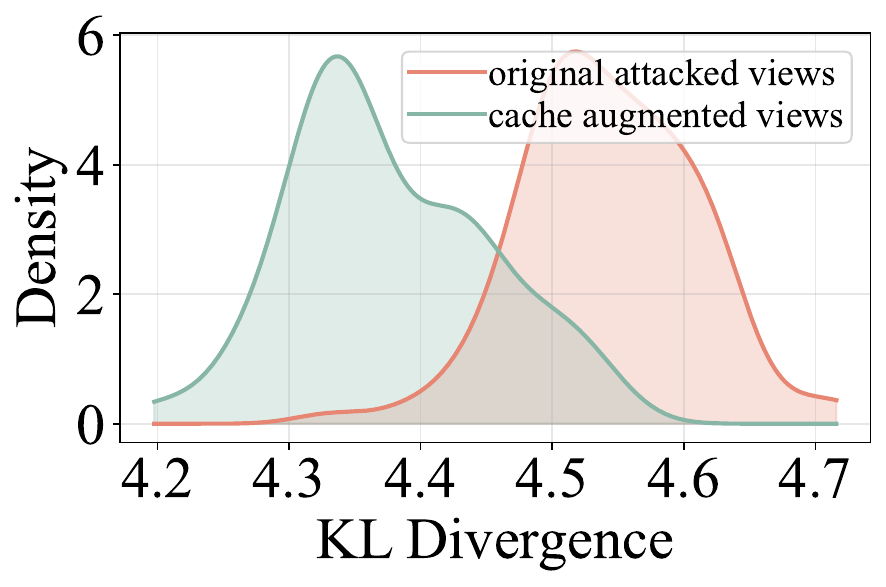}
        \caption{Caltech101}
        \label{fig:kl_density_Caltech101}
    \end{subfigure}
    
    \caption{Comparison of KL divergence distributions for semantic alignment between original visual features and augmented features from the cache mechanism. Kernel Density Estimation (KDE) curves are presented for (a) DTD and (b) Caltech101 datasets. Lower KL values signify more deterministic vision-text alignment. } % 总标题
    \label{fig:ablation_alignment}
    \vspace{-5pt}
\end{figure}

%% file: Sections/appendix.tex
\section{Proof of Theorem 1}
\label{proof}
\decomAlign*
\begin{proof}
 Let $X \sim \mu$ and $Z \sim \nu$ be random vectors in $\mathbb{R}^d$ with mean vectors $\boldsymbol{\mu}_x, \boldsymbol{\mu}_z$ and covariance matrices $\boldsymbol{\Sigma}_x, \boldsymbol{\Sigma}_z$, respectively. The 2-Wasserstein distance is defined as the minimum expected transport cost over all valid joint couplings $\pi \in \Pi(\mu, \nu)$:
\begin{equation}
W_2^2(\mu, \nu) = \inf_{\pi \in \Pi(\mu, \nu)} \mathbb{E}_{(X, Z) \sim \pi} \big[ \|X - Z\|^2 \big].
\end{equation}

We first expand the squared Euclidean cost by centering the variables around their respective means. Let $\tilde{X} = X - \boldsymbol{\mu}_x$ and $\tilde{Z} = Z - \boldsymbol{\mu}_z$. The cost function can be rewritten as:
\begin{equation}
\begin{aligned}
\|X - Z\|^2 &= \|(\tilde{X} - \tilde{Z}) + (\boldsymbol{\mu}_x - \boldsymbol{\mu}_z)\|^2 \\
&= \|\boldsymbol{\mu}_x - \boldsymbol{\mu}_z\|^2 + \|\tilde{X}\|^2 + \|\tilde{Z}\|^2 + 2(\boldsymbol{\mu}_x - \boldsymbol{\mu}_z)^\top(\tilde{X} - \tilde{Z}) - 2\tilde{X}^\top \tilde{Z}.
\end{aligned}
\end{equation}
Taking the expectation $\mathbb{E}_\pi[\cdot]$, the linear terms vanish because the variables are centered (i.e., $\mathbb{E}[\tilde{X}] = \mathbb{E}[\tilde{Z}] = 0$). Utilizing the identity $\mathbb{E}[\|\tilde{X}\|^2] = \operatorname{Tr}(\boldsymbol{\Sigma}_x)$, the expected cost simplifies to:
\begin{equation}
\label{eq:decomp_raw}
\mathbb{E}_\pi\big[\|X - Z\|^2\big] = \|\boldsymbol{\mu}_x - \boldsymbol{\mu}_z\|^2 + \operatorname{Tr}(\boldsymbol{\Sigma}_x) + \operatorname{Tr}(\boldsymbol{\Sigma}_z) - 2 \mathbb{E}_\pi \big[\tilde{X}^\top \tilde{Z}\big].
\end{equation}

To find $W_2^2$, we minimize Eq.~\eqref{eq:decomp_raw} over the coupling $\pi$. Since the mean difference and trace terms are constants independent of $\pi$, the problem reduces to maximizing the correlation term $\mathbb{E}_\pi [\tilde{X}^\top \tilde{Z}]$. For the family of elliptical distributions (e.g., Gaussians), it is a known result (Gelbrich, 1990) that the optimal coupling yields:
\begin{equation}
\sup_{\pi \in \Pi} \mathbb{E}_\pi \big[\tilde{X}^\top \tilde{Z}\big] = \operatorname{Tr}\left( (\boldsymbol{\Sigma}_x^{1/2} \boldsymbol{\Sigma}_z \boldsymbol{\Sigma}_x^{1/2})^{1/2} \right).
\end{equation}
Substituting this optimal correlation back into Eq.~\eqref{eq:decomp_raw}, we obtain:
\begin{equation}
\begin{aligned}
W_2^2(\mu, \nu) &= \|\boldsymbol{\mu}_x - \boldsymbol{\mu}_z\|^2 + \operatorname{Tr}(\boldsymbol{\Sigma}_x) + \operatorname{Tr}(\boldsymbol{\Sigma}_z) - 2\operatorname{Tr}\left( (\boldsymbol{\Sigma}_x^{1/2} \boldsymbol{\Sigma}_z \boldsymbol{\Sigma}_x^{1/2})^{1/2} \right) \\
&= \underbrace{\|\boldsymbol{\mu}_x - \boldsymbol{\mu}_z\|^2}_{\mathcal{L}_{\mathrm{mean}}} + \underbrace{\operatorname{Tr}\left( \boldsymbol{\Sigma}_x + \boldsymbol{\Sigma}_z - 2(\boldsymbol{\Sigma}_x^{1/2} \boldsymbol{\Sigma}_z \boldsymbol{\Sigma}_x^{1/2})^{1/2} \right)}_{\mathfrak{B}^2(\boldsymbol{\Sigma}_x, \boldsymbol{\Sigma}_z)}.
\end{aligned}
\end{equation}
The first term corresponds to the centroid distance ($\mathcal{L}_{\mathrm{mean}}$), and the second term is the squared Bures metric ($\mathfrak{B}^2$), representing the structural variance cost. For general distributions, this expression serves as a tight lower bound, confirming that minimizing $W_2^2$ inherently constrains both the first-order (mean) and second-order (variance) geometric moments. \hfill $\square$
\end{proof}

\clearpage
\section{Implementation Details}
\label{app:exp_detail}
For fair comparison, all approaches use the same pre-trained CLIP backbone and identical AugMix-based test-time augmentations, without external data, foundation models, or additional knowledge. We report average accuracy on clean samples and adversarial accuracy under PGD attacks with different perturbation budgets for default. Adversarial examples are generated on the original CLIP model, reflecting a realistic threat model.

For most experiments, we adopt a default setting that utilizes class descriptors and sets the subset size $|\mathcal{S}|=64$. However, to accommodate the distinct characteristics of specific benchmarks, we adjust these parameters for several datasets as follows: 1) For EuroSAT, we disable descriptors for the ViT-B/16 backbone and set $|\mathcal{S}|=32$ for both ViT-B/16 and ViT-B/32; 2) For ImageNet-A and ImageNet-R, the size of subset $|\mathcal{S}|$ is 32; 3) For ImageNet-Sketch, descriptors are disabled, and the subset size $|\mathcal{S}|$ is 32. For all other datasets not explicitly mentioned, the default configuration is maintained. These refinements are designed to better align domain-specific semantic features with the test-time adaptation process.

\section{Additional Results}
\subsection{Results on ImageNet and ImageNet-OOD datasets.}
\label{app_Imagenet}
%对于Imagenet以及他的四个变体数据集，我们同样进行了实验测试，结果见表2。实验结果表明，RITA 在处理大规模通用视觉任务时依然保持了显著的鲁棒性优势。在 ImageNet 上，RITA 在两种架构下均取得了最优或次优的对抗鲁棒性，例如在 ViT-B/16 上达到了 55.1% ，相比原始 CLIP 的 0.6% 有了质的飞跃。在更具挑战性的 OOD 变体中，RITA 的表现同样出色，其 OOD 平均对抗精度在两个骨架上分别达到 39.3%和46.2%，持续超越了 TPT、MTA 以及 R-TPT 等竞争对手。这些结果充分验证了 RITA 不仅在细粒度任务上表现卓越，在面对大规模通用场景及多样化的分布偏移时，依然能提供稳健的鲁棒性保护。
Table~\ref{tab:res_OOD} presents the performance comparison across ImageNet and its four Out-of-Distribution (OOD) variants. The results demonstrate that $\ours$ maintains a significant robustness advantage even when addressing large-scale general visual tasks. On the standard ImageNet, $\ours$ achieves state-of-the-art or competitive accuracies across both architectures under both settings; for instance, it reaches 55.1\% robustness on ViT-B/16, a substantial leap from the vanilla CLIP's 0.6\%. $\ours$'s performance is equally compelling on the more challenging OOD variants, where its average OOD robust accuracy reaches 39.3\% and 46.2\% in two backbones, outperforming other methods. These findings validate that $\ours$ not only excels in fine-grained tasks but also provides robust protection against diverse adversarial threats in large-scale general scenarios and under various distribution shifts.

\input{Tables/res_OOD}

\subsection{Evaluation under enhanced adversarial attacks.}
% \noindent{\textbf{Robustness evaluation under enhanced adversarial attacks.}}
%表7展示了在更严苛的攻击强度（$\epsilon=4.0$）下，不同测试时自适应方法在八个细粒度数据集上的表现。实验结果表明，随着攻击强度的增加，原始CLIP的精度已经接近于0，而 RITA 表现出卓越的抗干扰能力。具体而言，在两种骨干网络下，RITA的鲁棒精度均为最优，分别为32.9% 和 35.0%。这进一步证实了RITA 在保持高分类精度的同时，具备极强的预测稳定性，在应对现实世界中复杂且强烈的对抗风险时具有显著的实用价值。
Table~\ref{tab:ablation_attack_epsilon} presents the performance of various test-time adaptation methods across eight fine-grained datasets under a more stringent adversarial constraint, where $\epsilon$ is set to 4.0. The experimental results indicate that as the attack intensity increases, the accuracy of the vanilla CLIP drops nearly to zero, whereas $\ours$ demonstrates exceptional interference resistance. Specifically, our method achieves state-of-the-art results across both backbones, reaching 32.9\% and 35.0\% in average robust accuracy, respectively. These findings further confirm that $\ours$ maintains high classification precision while exhibiting remarkable predictive stability, thereby validating its substantial practical value in mitigating complex and intense adversarial risks in real-world scenarios.
\input{Tables/ablation_attack_epsilon}

\FloatBarrier
\subsection{Analysis on an alternative CLIP backbone.}
% \noindent{\textbf{Robustness evaluation on an alternative backbone.}}
%在Table x 中，我们进一步评估了 $\ours$ 在以 RN50 作为视觉骨干网络时的性能，以验证其在不同架构下的普适性。实验结果显示，在 $\epsilon=1.0$ 的对抗约束下，$\ours$ 在八个细粒度分类数据集上均表现出卓越的鲁棒性。值得注意的是，尽管 RN50 的基础表征能力弱于 ViT 系列，但 $\ours$ 依然能在保持高清洁精度的同时，超越了所有先进自适应方法。这一结果有力地证明了 $\ours$ 具有跨架构的鲁棒性增益，能够有效缓解不同视觉编码器在面对对抗攻击时的脆弱性。
In Table~\ref{tab:app_RN50}, we further evaluate the performance of $\ours$ using RN50 as the vision backbone to verify its generalizability across different architectures. The experimental results demonstrate that $\ours$ exhibits superior robustness across all eight fine-grained classification datasets. Notably, despite RN50 having a relatively weaker baseline representation capability compared to the ViT series, $\ours$ consistently outperforms other adaptation methods, while maintaining high clean accuracy. These findings provide strong evidence that our method delivers cross-architecture robustness gains and effectively mitigates the vulnerability of diverse vision encoders to adversarial attacks.
\input{Tables/appendix_RN50}

\subsection{Robustness evaluation under other attacks.}
\input{Tables/ablation_diff_atks}
We conduct experiments under various adversarial attack protocols, including adaptive attack that is aware of augmentation strategies, AutoAttack (AA), and FGSM. As reported in Table~\ref{tab:ablation_diff_atks}, $\ours$ consistently achieves the highest robust accuracy across all eight datasets and all attack types compared to the CLIP and R-TPT baselines. Specifically, under the more rigorous AutoAttack, $\ours$ maintains an average robust accuracy of 55.4\%, outperforming R-TPT by 2.8\%. Notably, $\ours$ shows significant gains on challenging datasets like DTD and EuroSAT across all attack settings. These results demonstrate that the defensive capability of $\ours$ is not tailored to a specific attack but generalizes well to diverse adversarial threats, confirming its potential for securing VLMs in various hostile environments.

\subsection{Generalization to Other VLM Backbones.}
\label{app_other_backbone}
\input{Tables/ablation_diff_backbones}
To further validate the architectural agnosticity of $\ours$, we extend our evaluation to other VLMs, including OpenCLIP and EVA-CLIP. As shown in Table~\ref{tab:ablation_diff_backbones}, we expanded $\ours$ to 10 datasets by incorporating SUN397 and Food101. The results consistently demonstrate that $\ours$ significantly boosts adversarial robustness across all architectures. For instance, when applied to OpenCLIP, $\ours$ improves the average robust accuracy from 2.4\% to 54.4\%. On the newly added SUN397 and Food101 datasets, $\ours$ achieves substantial gains, reaching robust accuracies of 54.0\% and 60.7\% for OpenCLIP, and 59.2\% and 62.4\% for EVA-CLIP, respectively. These findings underscore that the robustness gains of $\ours$ are consistent across diverse VLM backbones and broader dataset distributions, reinforcing its effectiveness as a general test-time adaptation framework.

\input{Figures/appendix_kl}

\section{Verification of the importance of the cache mechanism}
\subsection{Ablation study of the dynamic cache.}
\label{app_cache}
%表 x 展示了 RITA 主要组件的消融实验结果，重点验证了缓存机制（$d_{cOT}$）对鲁棒推理的贡献。实验结果表明，虽然仅使用缓存机制时，由于缺乏实时特征对齐，细粒度数据集和 ImageNet 的清洁精度分别下降至 39.7% 和 20.2%，但其对抗鲁棒性指标（分别为 27.1% 和 29.6%）仍显著优于原始 CLIP 基准（4.7% 和 3.2%）。这有力地证明了缓存中存储的历史先验即使在没有实时对齐的情况下，也能为极端对抗环境下的预测提供关键的参考信息。更重要的是，当缓存机制与最优传输模块（$d_{OT}$）协同工作时，模型在所有指标上均达到最优表现，尤其在细粒度数据集上将鲁棒性从单一模块的 49.4% 进一步提升至 55.0%。这充分说明缓存机制能够为多模态对齐提供不可或缺的语义补充，两者共同构建了一个更为稳健的推理框架。
In Table~\ref{tab:ablation_module}, we conduct an ablation study to specifically validate the significance of the cache mechanism $d_{cOT}$ within the $\ours$ framework. The results indicate that while utilizing the cache mechanism in isolation yields lower standard accuracy (39.7\% on fine-grained and 20.2\% on ImageNet) due to the absence of real-time alignment, it consistently outperforms the vanilla CLIP baseline in robustness metrics, achieving 27.1\% and 29.6\% respectively, compared to CLIP's 4.7\% and 3.2\%. This evidence underscores that the historical priors preserved in the cache serve as an essential reference for stabilizing predictions under adversarial perturbations. Most importantly, the synergy between the cache mechanism and the optimal transport module $d_{OT}$ leads to peak performance across all metrics, notably boosting fine-grained robustness from 49.4\% to 55.0\%. This further demonstrates that the cache provides critical semantic supplementation to the multimodal alignment, establishing a more robust and historically-aware inference framework.
\input{Tables/ablation_module}

\FloatBarrier
\subsection{Semantic alignment analysis of the cache mechanism.}
\label{app_alignment}
%为了验证 Cache 机制在分类任务中的有效性，我们提取了 Caltech101 和DTD数据集推理后的增强视图特征，并从“视觉-文本”对齐的角度将其与原始无增强视觉特征进行对比。具体而言，对于每个类别，我们首先计算文本原型上的类条件分配分布 $\tilde{\mathbb{P}}$。该概率分布通过对视觉特征与文本嵌入之间的欧氏距离平方执行 Softmax 操作获得，反映了视觉特征在文本语义空间中的解释方式。随后，我们测量该经验分布 $\tilde{\mathbb{P}}$ 与理想单热点（One-hot）目标分布 $\tilde{\mathbb{P}}_{\mathrm{gt}}$之间的 Kullback-Leibler (KL) 散度。较低的 KL 值意味着更强的视觉-文本对齐和更低的语义歧义。
%我们在图 X 中展示了 Caltech101 和 DTD 数据集中所有类别的 KL 散度散点分布。图中每个点代表一个特定类别的视觉-文本对齐质量，其中红色点代表原始对抗视图，绿色点代表低熵增强视图（即 Cache 机制所使用的特征）。原始视图（红色）的 KL 散度波动较大且维持在较高水平，反映出对抗扰动导致模型在不同类别上均产生了严重的预测偏差。引入 Cache 机制后，绿色点的分布不仅下移，且其离散程度有所降低，说明增强视图通过聚合历史先验，有效地校准了受损的特征表示，使模型能够回归到更接近理想单热点分布（One-hot distribution）的状态。即便在原始 KL 散度极高的困难类别上，绿色特征依然展现出大幅度的对齐增益。这种跨类别的稳定性是 RITA 框架在面对未知对抗攻击时展现出卓越鲁棒性的微观基础。
To demonstrate the effectiveness of the cache mechanism, we extract the augmented view features from the DTD and Caltech101 datasets and compare them with the original unaugmented visual features from a "vision-text" alignment perspective. Specifically, for each class, we first compute the class-conditional assignment distribution over all text prototypes, which reflects how the visual features of each class are semantically aligned with the textual features. Subsequently, we measure the KL divergence between this empirical distribution and an ideal one-hot target distribution that assigns all probability mass to the ground-truth class. A lower KL value indicates stronger vision-text alignment.

As illustrated in Figure~\ref{fig:appendix_kl}, we present the scatter plots of KL divergence for all classes in DTD and Caltech101. Each point represents the vision-text alignment quality of a specific category, where red dots denote original adversarial views and green dots represent selected augmented views utilized by our cache mechanism. The KL divergence of original views exhibits high variance and remains at an elevated level, reflecting severe adversarial bias. Upon introducing the cache mechanism, the green dots show both a downward shift and reduced dispersion. This indicates that the augmented views effectively calibrate the corrupted feature representations by aggregating historical priors, pulling the model closer to the ideal one-hot distribution. Even for categories where the original KL divergence is particularly high, the augmented features still achieve significant alignment gains. This cross-category consistency provides a granular foundation for the superior robustness of the $\ours$ framework when encountering diverse adversarial attacks.

\input{Tables/algorithm}

\section{Algorithm for RITA}
\label{app_algorithm}
Algorithm \ref{alg:rita} summarizes the $\ours$ framework. $\ours$ extracts features from augmented views, aligns visual-textual distributions via optimal transport, and maintains a dynamic cache for progressive refinement.

%% file: Tables/res_OOD.tex
\begin{table*}[!t]
\centering
\tabstyle{5pt}
\caption{Results (\%) of various adaptation methods on ImageNet and ImageNet-OOD datasets with $\epsilon$ = 1.0. OOD Avg. refers to the average results among four ImageNet-OOD datasets.}
\begin{tabular}{c| l|cc|cc|cc|cc|cc|cc}
\toprule
\multirow{2}{*}{} &
\multirow{2}{*}{Method} &
\multicolumn{2}{c|}{ImageNet} &
\multicolumn{2}{c|}{ImageNet-A} &
\multicolumn{2}{c|}{ImageNet-V2} &
\multicolumn{2}{c|}{ImageNet-R} &
\multicolumn{2}{c|}{ImageNet-S} &
\multicolumn{2}{c}{OOD Avg.} \\
 &  &
Acc. & Rob. &
Acc. & Rob. &
Acc. & Rob. &
Acc. & Rob. &
Acc. & Rob. &
Acc. & Rob. \\
\midrule
\multirow{7}{*}{\rotatebox{90}{ViT-B/32}}
& CLIP
& 62.0 & 0.7 & 29.5 & 0.1 & 54.7 & 1.5 & 66.2 & 6.9 & 40.8 & 4.5 & 47.8 & 3.2 \\

& Ensemble
& 64.4 & \underline{52.3} & 34.1 & 21.4 & 58.1 & \underline{45.6} & 64.5 & 55.2 & 39.2 & 31.2 & 49.0 & 38.3 \\

& TPT
& 63.6 & 36.6 & 34.5 & 9.3 & 56.9 & 30.4 & \underline{69.1} & 49.4 & 41.6 & 30.4 & 50.5 & 31.2 \\

& C-TPT
& 63.5 & 33.4 & 30.5 & 7.6 & 55.9 & 27.4 & 67.0 & 45.0 & \underline{41.8} & 30.1 & 48.8 & 28.7 \\

& MTA
& \bestclean{64.9} & 40.1 & \bestclean{37.7} & 11.1 & \underline{58.3} & 33.2 & \bestclean{70.3} & 52.3 & \bestclean{43.4} & \underline{31.5} & \bestclean{52.4} & 32.0 \\

& R-TPT
& 64.4 & 52.1 & \underline{36.9} & \underline{21.9} & 58.0 & 45.5 & 67.5 & \underline{55.8} & 41.7 & 31.2 & \underline{51.0} & \underline{38.6} \\

& \cellcolor{rowgreen}\ours
& \cellcolor{rowgreen}\underline{64.8} & \cellcolor{rowgreen}\bestclean{52.7} & \cellcolor{rowgreen}35.4 & \cellcolor{rowgreen}\bestclean{22.5} & \cellcolor{rowgreen}\bestclean{58.5} & \cellcolor{rowgreen}\bestclean{45.9} & \cellcolor{rowgreen}65.7 & \cellcolor{rowgreen}\bestclean{56.3} & \cellcolor{rowgreen}40.8 & \cellcolor{rowgreen}\bestclean{32.3} & \cellcolor{rowgreen}50.1& \cellcolor{rowgreen}\bestclean{39.3} \\ \midrule

\multirow{7}{*}{\rotatebox{90}{ViT-B/16}}
& CLIP
& 66.7 & 0.6 & 47.7 & 0.1 & 60.8 & 0.2 & 73.9 & 3.5 & 46.1 & 2.2 & 57.1 & 1.5 \\

& Ensemble
& 68.8 & \underline{54.4} & 55.8 & 33.6 & 62.8 & 47.4 & 72.9 & 62.7 & 46.5 & 35.1 & 59.5 & 44.7 \\

& TPT
& 68.9 & 42.4 & 54.7 & 14.9 & \bestclean{63.6} & 35.7 & \bestclean{77.1} & 57.3 & \underline{47.9} & 35.6 & 60.8 & 37.1 \\

& C-TPT
& 68.1 & 38.0 & 49.7 & 11.3 & 61.9 & 31.4 & 74.8 & 51.9 & 47.2 & 34.4 & 58.4 & 3.4 \\

& MTA 
& \underline{69.0} & 44.4 & \bestclean{57.3} & 17.5 & 63.4 & 37.2 & \underline{76.9} & 58.9 & \bestclean{48.4} & 35.8 & \bestclean{61.5} & 38.7  \\

& R-TPT
& \bestclean{69.1} & \underline{54.4} & \underline{57.2} & \underline{34.7} & \underline{63.5} & \underline{48.0} & 75.5 & \underline{63.7} & 47.7 & \underline{36.5} & \underline{60.9} & \underline{45.7}  \\

& \cellcolor{rowgreen}\ours
& \cellcolor{rowgreen}\bestclean{69.1} & \cellcolor{rowgreen}\bestclean{55.1} & \cellcolor{rowgreen}55.8 & \cellcolor{rowgreen}\bestclean{35.0} & \cellcolor{rowgreen}63.2 & \cellcolor{rowgreen}\bestclean{48.4} & \cellcolor{rowgreen}74.0 & \cellcolor{rowgreen}\bestclean{64.1} & \cellcolor{rowgreen}47.1 & \cellcolor{rowgreen}\bestclean{37.2} & \cellcolor{rowgreen}60.0 & \cellcolor{rowgreen}\bestclean{46.2}  \\ 
\bottomrule
\end{tabular}
\label{tab:res_OOD}
\end{table*}

%% file: Tables/ablation_attack_epsilon.tex
\begin{table*}[ht]
\centering
\tabstyle{2.5pt}
\caption{Results (\%) of adaptation methods on fine-grained classification datasets with $\epsilon$ = 4.0. 
% \bestclean{Bold} and \underline{underlined} denote best and second-best. 
}
\begin{tabular}{c| l|cc|cc|cc|cc|cc|cc|cc|cc|cc}
\toprule
\multirow{2}{*}{} & 
\multirow{2}{*}{Method} & 
\multicolumn{2}{c|}{Caltech101} & 
\multicolumn{2}{c|}{Pets} & 
\multicolumn{2}{c|}{Cars} & 
\multicolumn{2}{c|}{Flower102} & 
\multicolumn{2}{c|}{Aircraft} & 
\multicolumn{2}{c|}{DTD} & 
\multicolumn{2}{c|}{EuroSAT} & 
\multicolumn{2}{c|}{UCF101} & 
\multicolumn{2}{c}{Avg.} \\
 &  & 
Acc. & Rob. & 
Acc. & Rob. & 
Acc. & Rob. & 
Acc. & Rob. & 
Acc. & Rob. & 
Acc. & Rob. & 
Acc. & Rob. & 
Acc. & Rob. & 
Acc. & Rob. \\
\midrule
\multirow{7}{*}{\rotatebox{90}{ViT-B/32}} 
& CLIP
& 90.9 & 2.5 & 83.0 & 0.0 & 49.7 & 0.0 & 65.8 & 0.0 & 18.3 & 0.0 & 40.8 & 0.2 & 18.6 & 0.0 & 62.1 & 0.0 & 53.6 & 0.3  \\

& Ensemble 
& 91.6 & 73.7 & 85.0 & \underline{47.0} & 57.8 & 18.5 & \underline{67.4} & \underline{34.8} & 20.1 & \underline{7.2} & \underline{46.1} & \underline{29.2} & 32.5 & \underline{7.4} & 61.6 & \underline{37.8} & \underline{57.8} & \underline{32.0}  \\

& TPT
& 91.4 & 60.0 & 84.1 & 30.3 & 62.9 & 16.4 & 63.8 & 28.5 & 19.0 & 3.9 & 42.2 & 19.9 & \bestclean{35.1} & 6.5 & 62.3 & 21.8 & 57.6 & 23.4  \\

& C-TPT
& \underline{91.8} & 53.7 & 84.9 & 21.5 & 60.8 & 9.2 & 65.9 & 22.1 & 17.7 & 2.4 & 44.3 & 16.4 & \underline{34.7} & 5.7 & 62.6 & 17.8 & \underline{57.8} & 18.6  \\

& MTA
& \underline{91.8} & 73.9 & \underline{85.8} & 45.8 & \bestclean{64.1} & 19.2 & 64.8 & 34.2 & \bestclean{20.4} & 6.1 & 44.0 & 22.0 & 34.5 & 5.3 & \bestclean{63.6} & 33.1 & \bestclean{58.6} & 29.9  \\

& R-TPT 
& 90.6 & \underline{74.6} & 84.5 & 44.5 & \underline{63.1} & \bestclean{20.5} & 62.6 & 34.1 & 19.1 & 6.6 & 42.1 & 27.3 & 32.0 & 7.1 & \underline{62.8} & 37.1 & 57.1 & 31.4  \\

& \cellcolor{rowgreen}\ours
& \cellcolor{rowgreen}\bestclean{92.3} & \cellcolor{rowgreen}\bestclean{74.8} & \cellcolor{rowgreen}\bestclean{85.9} & \cellcolor{rowgreen}\bestclean{47.4} & \cellcolor{rowgreen}59.6 & \cellcolor{rowgreen}\underline{20.1} & \cellcolor{rowgreen}\bestclean{68.7} & \cellcolor{rowgreen}\bestclean{35.6} & \cellcolor{rowgreen}\underline{20.2} & \cellcolor{rowgreen}\bestclean{7.7} & \cellcolor{rowgreen}\bestclean{46.2} & \cellcolor{rowgreen}\bestclean{30.2} & \cellcolor{rowgreen}33.4 & \cellcolor{rowgreen}\bestclean{8.5} & \cellcolor{rowgreen}\underline{62.8} & \cellcolor{rowgreen}\bestclean{39.1} & \cellcolor{rowgreen}\bestclean{58.6} & \cellcolor{rowgreen}\bestclean{32.9}  \\ \midrule

\multirow{7}{*}{\rotatebox{90}{ViT-B/16}} 
& CLIP 
& 85.9 & 0.7 & 83.5 & 0.0 & 55.7 & 0.0 & 61.7 & 0.0 & 15.7 & 0.0 & 40.4 & 0.0 & 23.7 & 0.0 & 58.9 & 0.0 & 53.2 & 0.1  \\

& Ensemble 
& 92.1 & 76.8 & \underline{88.7} & \underline{47.9} & 63.2 & 22.4 & \underline{70.8} & \underline{37.6} & \underline{25.9} & 10.2 & \underline{50.9} & \underline{33.2} & 32.9 & \underline{6.6} & 64.6 & \underline{35.6} & 61.1 & \underline{33.8}  \\

& TPT
& \underline{94.1} & 60.6 & 87.4 & 31.0 & 66.5 & 13.8 & 66.1 & 23.7 & 23.4 & 4.4 & 45.9 & 17.4 & \bestclean{42.6} & 4.6 & \bestclean{67.9} & 20.3 & 61.7 & 21.9  \\

& C-TPT 
& 93.9 & 49.6 & 88.2 & 21.1 & 65.8 & 9.2 & 69.6 & 17.2 & 23.9 & 2.0 & 45.9 & 12.7 & 42.3 & 5.2 & 65.6 & 14.2 & \underline{61.9} & 16.4  \\

& MTA 
& \bestclean{94.3} & 73.6 & 88.0 & 51.2 & \bestclean{67.7} & \bestclean{25.7} & 65.0 & 31.7 & 24.0 & 7.4 & 46.5 & 21.5 & \underline{42.5} & 6.5 & \underline{67.5} & 30.9 & \underline{61.9} & 31.0  \\

& R-TPT 
& 93.7 & \underline{78.3} & 87.2 & 45.6 & \underline{67.0} & 23.9 & 68.7 & 34.8 & 23.9 & \underline{10.5} & 46.4 & 30.4 & 34.7 & 6.3 & 67.2 & 35.2 & 61.1 & 33.1  \\

& \cellcolor{rowgreen}\ours
& \cellcolor{rowgreen}93.8 & \cellcolor{rowgreen}\bestclean{78.5} & \cellcolor{rowgreen}\bestclean{89.8} & \cellcolor{rowgreen}\bestclean{48.1} & \cellcolor{rowgreen}64.2 & \cellcolor{rowgreen}\underline{24.2} & \cellcolor{rowgreen}\bestclean{71.6} & \cellcolor{rowgreen}\bestclean{38.4} & \cellcolor{rowgreen}\bestclean{26.2} & \cellcolor{rowgreen}\bestclean{11.3} & \cellcolor{rowgreen}\bestclean{51.5} & \cellcolor{rowgreen}\bestclean{34.6} & \cellcolor{rowgreen}33.4 & \cellcolor{rowgreen}\bestclean{7.9} & \cellcolor{rowgreen}65.5 & \cellcolor{rowgreen}\bestclean{37.2} & \cellcolor{rowgreen}\bestclean{62.0} & \cellcolor{rowgreen}\bestclean{35.0}  \\

\bottomrule
\end{tabular}
\label{tab:ablation_attack_epsilon}
\end{table*}

%% file: Tables/appendix_RN50.tex
\begin{table*}[!t]
\centering
\tabstyle{2.5pt}
\caption{Results (\%) of adaptation methods on fine-grained classification datasets using RN50 with $\epsilon$ set to 1.0.
% \bestclean{Bold} and \underline{underlined} entries indicate the best and second-best results, respectively.
}
\begin{tabular}{l|cc|cc|cc|cc|cc|cc|cc|cc|cc}
\toprule
Method & 
\multicolumn{2}{c|}{Caltech101} & 
\multicolumn{2}{c|}{Pets} & 
\multicolumn{2}{c|}{Cars} & 
\multicolumn{2}{c|}{Flower102} & 
\multicolumn{2}{c|}{Aircraft} & 
\multicolumn{2}{c|}{DTD} & 
\multicolumn{2}{c|}{EuroSAT} & 
\multicolumn{2}{c|}{UCF101} & 
\multicolumn{2}{c}{Avg.} \\
 & 
Acc. & Rob. & 
Acc. & Rob. & 
Acc. & Rob. & 
Acc. & Rob. & 
Acc. & Rob. & 
Acc. & Rob. & 
Acc. & Rob. & 
Acc. & Rob. & 
Acc. & Rob. \\
\midrule

CLIP 
& 84.9 & 2.6 & 83.5 & 0.0 & 53.7 & 0.0 & 61.7 & 0.0 & 14.7 & 0.0 & 40.4 & 0.8 & 18.7 & 0.0 & 57.5 & 0.0 & 51.8 & 0.4 \\

Ensemble 
& 84.6 & \underline{78.2} & \underline{85.1} & \underline{75.4} & 52.6 & 38.4 & \underline{65.4} & \underline{56.2} & 15.7 & \underline{11.8} & \underline{43.1} & \underline{38.2} & 22.0 & 14.6 & 56.3 & \underline{49.5} & 53.1 & \underline{45.2} \\

TPT
& 85.1 & 7.0 & 84.7 & 0.1 & 54.4 & 0.0 & 62.1 & 0.0 & 15.3 & 5.2 & 42.4 & 4.3 & \underline{22.4} & 0.0 & \underline{60.2} & 0.3 & 53.3 & 2.1 \\

C-TPT
& 84.8 & 3.7 & 83.6 & 0.0 & 55.6 & 0.0 & 64.8 & 0.0 & \underline{16.7} & 7.2 & 41.5 & 1.3 & 22.2 & 0.0 & 60.1 & 0.1 & \underline{53.6} & 1.5 \\

MTA
& 85.3 & 65.9 & 84.8 & 59.8 & \underline{55.7} & 17.8 & 61.0 & 31.5 & 15.9 & 10.3 & 40.3 & 18.8 & \bestclean{22.5} & 1.6 & \bestclean{60.6} & 31.3 & 53.2 & 29.6 \\

R-TPT
& \bestclean{86.7} & \underline{78.2} & 84.6 & 74.2 & \bestclean{56.1} & \underline{38.6} & 60.6 & 51.9 & 16.4 & \underline{11.8} & 41.3 & 33.5 & 21.2 & \underline{15.1} & 59.5 & 49.2 & 53.3 & 43.0 \\

\cellcolor{rowgreen}\ours
& \cellcolor{rowgreen}\underline{85.7} & \cellcolor{rowgreen}\bestclean{79.3}
& \cellcolor{rowgreen}\bestclean{86.3} & \cellcolor{rowgreen}\bestclean{77.4}
& \cellcolor{rowgreen}54.2 & \cellcolor{rowgreen}\bestclean{39.9}
& \cellcolor{rowgreen}\bestclean{66.2} & \cellcolor{rowgreen}\bestclean{56.9}
& \cellcolor{rowgreen}\bestclean{16.9} & \cellcolor{rowgreen}\bestclean{12.4}
& \cellcolor{rowgreen}\bestclean{43.8} & \cellcolor{rowgreen}\bestclean{39.5}
& \cellcolor{rowgreen}20.2 & \cellcolor{rowgreen}\bestclean{15.9}
& \cellcolor{rowgreen}58.8 & \cellcolor{rowgreen}\bestclean{50.0}
& \cellcolor{rowgreen}\bestclean{54.0} & \cellcolor{rowgreen}\bestclean{46.4} \\

\bottomrule
\end{tabular}
\label{tab:app_RN50}
\end{table*}

%% file: Tables/ablation_diff_atks.tex
\begin{table*}[t!]
\centering
\tabstyle{3pt}
\caption{Robust accuracy (\%) on fine-grained classification datasets under different attacks using ViT-B/16 with $\epsilon$ = 1.0. 
% \bestclean{Bold} and \underline{underlined} denote best and second-best.
}
\begin{tabular}{c| l | *{9}{>{\centering\arraybackslash}p{1.2cm}}}
\toprule
 & Method & Caltech101 & Pets & Cars & Flower102 & Aircraft & DTD & EuroSAT & UCF101 & Avg. \\
\midrule
\multirow{3}{*}{\rotatebox{90}{adaptive}} 
& CLIP  & 30.4 & 4.9 & 7.2 & 3.2 & 0.2 & 2.4 & 0.0 & 0.8 & 6.1 \\
& R-TPT & \underline{87.3} & \underline{76.3} & \underline{55.3} & \underline{60.8} & \underline{17.4} & \underline{35.3} & \underline{20.5} & \underline{52.2} & \underline{50.6} \\
& \cellcolor{rowgreen}\ours & \cellcolor{rowgreen}\textbf{88.2} & \cellcolor{rowgreen}\textbf{79.5} & \cellcolor{rowgreen}\textbf{57.8} & \cellcolor{rowgreen}\textbf{66.2} & \cellcolor{rowgreen}\textbf{19.1} & \cellcolor{rowgreen}\textbf{37.6} & \cellcolor{rowgreen}\textbf{23.8} & \cellcolor{rowgreen}\textbf{54.8} & \cellcolor{rowgreen}\textbf{53.4} \\ 
\midrule
\multirow{3}{*}{\rotatebox{90}{AA}} 
& CLIP  & 13.2 & 4.9 & 0.3 & 2.6 & 0.0 & 0.0 & 0.0 & 4.6 & 3.2 \\
& R-TPT & \underline{87.9} & \underline{78.0} & \underline{51.4} & \underline{59.4} & \underline{20.3} & \underline{41.5} & \underline{24.2} & \underline{58.2} & \underline{52.6} \\
& \cellcolor{rowgreen}\ours & \cellcolor{rowgreen}\textbf{89.8} & \cellcolor{rowgreen}\textbf{82.0} & \cellcolor{rowgreen}\textbf{53.7} & \cellcolor{rowgreen}\textbf{61.8} & \cellcolor{rowgreen}\textbf{22.6} & \cellcolor{rowgreen}\textbf{47.7} & \cellcolor{rowgreen}\textbf{26.0} & \cellcolor{rowgreen}\textbf{59.9} & \cellcolor{rowgreen}\textbf{55.4} \\ 
\midrule
\multirow{3}{*}{\rotatebox{90}{FGSM}} 
& CLIP  & 6.2 & 2.4 & 0.5 & 0.0 & 0.0 & 0.4 & 0.0 & 1.8 & 1.4 \\
& R-TPT & \underline{84.8} & \underline{73.6} & \underline{43.6} & \underline{54.3} & \underline{19.9} & \underline{36.2} & \underline{23.1} & \underline{50.3} & \underline{48.2} \\
& \cellcolor{rowgreen}\ours & \cellcolor{rowgreen}\textbf{85.9} & \cellcolor{rowgreen}\textbf{74.5} & \cellcolor{rowgreen}\textbf{44.2} & \cellcolor{rowgreen}\textbf{59.4} & \cellcolor{rowgreen}\textbf{21.7} & \cellcolor{rowgreen}\textbf{42.1} & \cellcolor{rowgreen}\textbf{24.4} & \cellcolor{rowgreen}\textbf{51.8} & \cellcolor{rowgreen}\textbf{50.5} \\
\bottomrule
\end{tabular}
\label{tab:ablation_diff_atks}
\end{table*}

%% file: Tables/ablation_diff_backbones.tex
\begin{table*}[t]
\caption{Classification accuracy (\%) on 10 datasets using EVA-CLIP and OpenCLIP backbones. }
\label{tab:ablation_diff_backbones}
\centering
\tabstyle{3pt}
\resizebox{\textwidth}{!}{%
\begin{tabular}{c|cc|cc|cc|cc|cc|cc|cc|cc|cc|cc|cc}
\toprule
\multirow{2}{*}{\rotatebox{0}{Method}} & \multicolumn{2}{c|}{\rotatebox{0}{Caltech101}} & \multicolumn{2}{c|}{\rotatebox{0}{Pets}} & \multicolumn{2}{c|}{\rotatebox{0}{Cars}} & \multicolumn{2}{c|}{\rotatebox{0}{Flower102}} & \multicolumn{2}{c|}{\rotatebox{0}{Aircraft}} & \multicolumn{2}{c|}{\rotatebox{0}{DTD}} & \multicolumn{2}{c|}{\rotatebox{0}{EuroSAT}} & \multicolumn{2}{c|}{\rotatebox{0}{UCF101}} & \multicolumn{2}{c|}{\rotatebox{0}{SUN397}} & \multicolumn{2}{c|}{\rotatebox{0}{Food101}} & \multicolumn{2}{c}{\rotatebox{0}{Avg.}}  \\ 
& Acc. & Rob. 
& Acc. & Rob. 
& Acc. & Rob.
& Acc. & Rob.
& Acc. & Rob.
& Acc. & Rob.
& Acc. & Rob.
& Acc. & Rob. 
& Acc. & Rob. 
& Acc. & Rob. 
& Acc. & Rob. \\
\midrule
OpenCLIP & 91.3 & 12.3 & 89.2 & 1.2 & 75.7 & 2.9 & 66.9 & 0.2 & 17.7 & 0.0 & 51.3 & 3.1 & 50.1 & 0.4 & 67.3 & 0.1 & 69.6 & 1.9 & 85.9 & 1.4 & 66.5 & 2.4 \\
+ \cellcolor{rowgreen}{\ours} & \cellcolor{rowgreen}\textbf{92.4} & \cellcolor{rowgreen}\textbf{89.9} & \cellcolor{rowgreen}\textbf{91.7} & \cellcolor{rowgreen}\textbf{78.2} & \cellcolor{rowgreen}\textbf{78.4} & \cellcolor{rowgreen}\textbf{50.2} & \cellcolor{rowgreen}\textbf{73.2} & \cellcolor{rowgreen}\textbf{62.4} & \cellcolor{rowgreen}\textbf{25.1} & \cellcolor{rowgreen}\textbf{19.3} & \cellcolor{rowgreen}\textbf{55.9} & \cellcolor{rowgreen}\textbf{46.3} & \cellcolor{rowgreen}\textbf{51.5} & \cellcolor{rowgreen}\textbf{29.9} & \cellcolor{rowgreen}\textbf{69.6} & \cellcolor{rowgreen}\textbf{53.5} & \cellcolor{rowgreen}\textbf{75.2} & \cellcolor{rowgreen}\textbf{54.0} & \cellcolor{rowgreen}\textbf{88.3} & \cellcolor{rowgreen}\textbf{60.7} & \cellcolor{rowgreen}\textbf{70.1} & \cellcolor{rowgreen}\textbf{54.4} \\ 
\midrule
EVA-CLIP & 86.3 & 5.2 & 92.2 & 0.3 & 78.6 & 4.5 & 75.9 & 1.2 & 24.8 & 0.0 & 53.1 & 1.7 & 67.0 & 0.5 & 63.2 & 0.0 & 79.7 & 4.2 & 89.4 & 0.9 & 71.0 & 1.6 \\
+ \cellcolor{rowgreen}{\ours} & \cellcolor{rowgreen}\textbf{87.1} & \cellcolor{rowgreen}\textbf{84.6} & \cellcolor{rowgreen}\textbf{93.1} & \cellcolor{rowgreen}\textbf{80.4} & \cellcolor{rowgreen}\textbf{79.5} & \cellcolor{rowgreen}\textbf{48.2} & \cellcolor{rowgreen}\textbf{78.6} & \cellcolor{rowgreen}\textbf{64.8} & \cellcolor{rowgreen}\textbf{27.5} & \cellcolor{rowgreen}\textbf{17.5} & \cellcolor{rowgreen}\textbf{56.2} & \cellcolor{rowgreen}\textbf{41.4} & \cellcolor{rowgreen}\textbf{69.8} & \cellcolor{rowgreen}\textbf{38.2} & \cellcolor{rowgreen}\textbf{64.9} & \cellcolor{rowgreen}\textbf{50.8} & \cellcolor{rowgreen}\textbf{82.3} & \cellcolor{rowgreen}\textbf{59.2} & \cellcolor{rowgreen}\textbf{91.2} & \cellcolor{rowgreen}\textbf{62.4} & \cellcolor{rowgreen}\textbf{73.0} & \cellcolor{rowgreen}\textbf{54.8} \\ 
\bottomrule
\end{tabular}%
}
\end{table*}

%% file: Figures/appendix_kl.tex
\begin{figure}[!t]
    \centering
    % 第一张子图
    \begin{subfigure}{0.4\columnwidth}
        \centering
        \includegraphics[width=\linewidth]{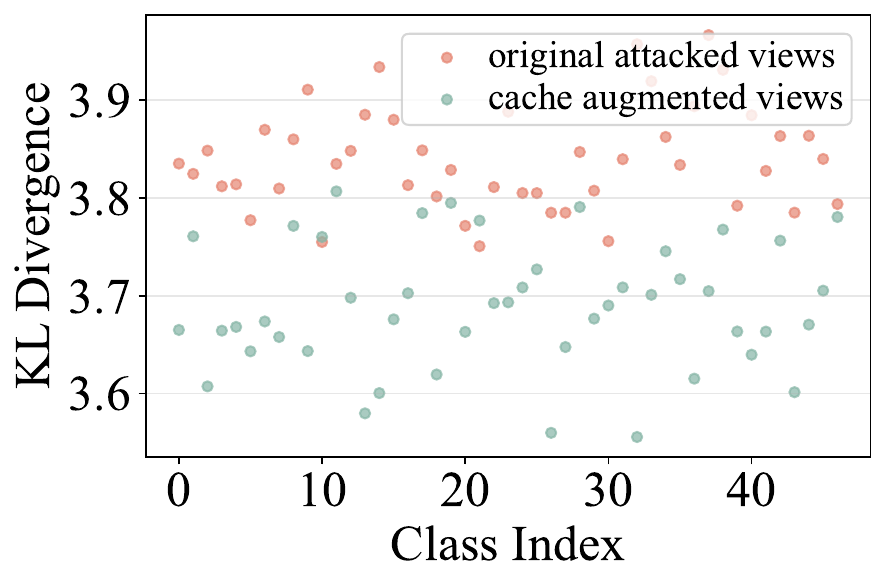}
        \caption{DTD}
        \label{fig:kl_scatter_DTD}
    \end{subfigure}
    \hfill % 在两图之间填充空白
    % 第二张子图
    \begin{subfigure}{0.4\columnwidth}
        \centering
        \includegraphics[width=\linewidth]{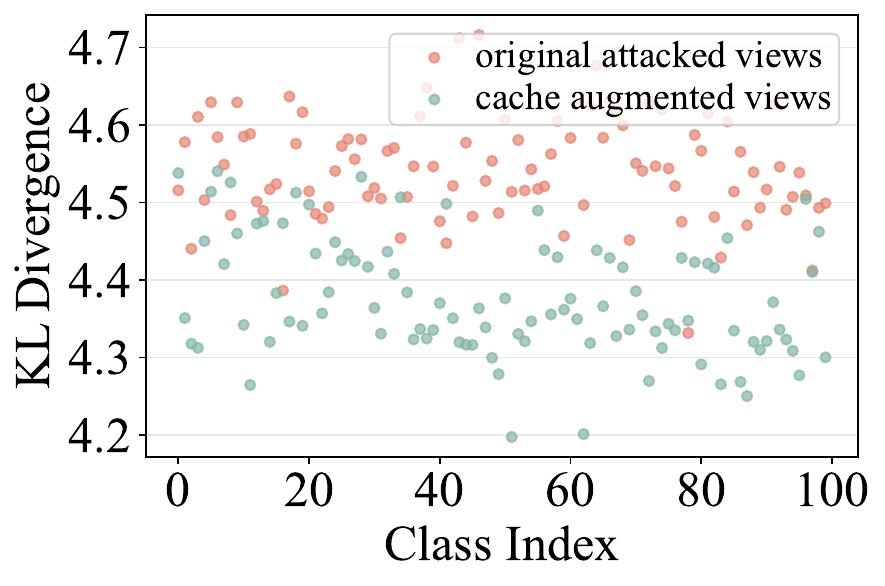}
        \caption{Caltech101}
        \label{fig:kl_scatter_Caltech101}
    \end{subfigure}
    
    \caption{KL divergence per class on (a) DTD and (b) Caltech101. Lower KL values signify superior vision-text alignment.} % 总标题
    \label{fig:appendix_kl}
\end{figure}

%% file: Tables/ablation_module.tex
    \begin{table}[!t]
        \centering
        \caption{ Main component analysis (\%) on fine-grained datasets and ImageNet dataset using ViT-B/32 with $\epsilon$ = 1.0.}
        % \vspace{-3mm}
        \centering
        % \resizebox{\linewidth}{!}
        % {
            \begin{tabular}{cc|cccc}
            \toprule
            \multirow{2}{*}{$d_{\mathrm{OT}}$} & \multirow{2}{*}{$d_{\mathrm{cOT}}$} & \multicolumn{2}{c}{Fine-grained}  & \multicolumn{2}{c}{ImageNet} \\
             &   & Acc. & Rob. & Acc. & Rob. \\
            \midrule
            \XSolidBrush & \XSolidBrush  & 53.6 & 4.7 & 47.8 & 3.2 \\
            \CheckmarkBold & \XSolidBrush  & 57.0 & 49.4 & 63.9 & 52.3 \\
            \XSolidBrush   & \CheckmarkBold & 39.7 & 27.1  & 20.2 & 29.6 \\
            \CheckmarkBold & \CheckmarkBold  & \bestclean{57.9} & \bestclean{55.0} & \bestclean{64.8}  & \bestclean{52.7} \\
            \bottomrule
            \end{tabular}
        % }
    \label{tab:ablation_module}
    \end{table}

% p(k \mid \hat{x}_t) \leftarrow p(k \mid \hat{x}_t) + p_\mathrm{cache}(k \mid \hat{x}_t).

%% file: Tables/algorithm.tex
\begin{algorithm}[t!]
    \vspace{-0.5em}
   \caption{RITA: Robust test-tIme prompT Adaptation}
   \label{alg:rita}
\begin{algorithmic}[1]
   \STATE \textbf{Input:} Test stream $\{\hat{x}_t\}$, encoders $\Phi_{\mathrm{img}}, \Phi_{\mathrm{text}}$, text prompts $\{z_k^{(m)}\}_{m=1}^M$, entropy threshold $\gamma$, max cache size per class $N_{k}$, cache weight $\alpha$. 
   \STATE \textbf{Initialize:} Empty cache $\{\hat{X}_k\}_{k=1}^K \leftarrow \emptyset$; extract text features $\mathbf{z}_k^m = \Phi_{\mathrm{text}}(z_k^{(m)})$ and construct text distributions $\mathbb{Q}_k = \frac{1}{M}\sum_{m=1}^M \delta_{\mathbf{z}_k^m}$.
   \FOR{each test image $\hat{x}_t$}
       \STATE Generate $N$ augmented views $\{\hat{x}_t^n\}_{n=1}^N$, extract visual features $\mathbf{x}_t^n = \Phi_{\mathrm{img}}(\hat{x}_t^n)$, and construct visual distribution $\mathbb{P}_t = \frac{1}{N}\sum_{n=1}^N \delta_{\mathbf{x}_t^n}$.
       
       \STATE \textbf{Cache Update:} For each view $\hat{x}_t^n$ with entropy $H(p_t^n) < \gamma$, get pseudo-label $\hat{k} = \argmax\limits_k p_t^n(k)$. Add $\mathbf{x}_t^n$ to $\hat{X}_{\hat{k}}$ (if $|\hat{X}_{\hat{k}}| \ge N_{k}$ and $H(p_t^n)$ is lower, replace the max-entropy sample).
       
       \FOR{class $k = 1$ \textbf{to} $K$}
           \STATE Compute global OT distance $d_{OT}(\mathbb{P}_t, \mathbb{Q}_k; C_{t,k})$ where $C_{t,k}(n,m) = 1 - \cos(\mathbf{x}_t^n, \mathbf{z}_k^m)$.
           \STATE \textbf{If} $\hat{X}_k \neq \emptyset$, align features $\tilde{X}_k = \hat{X}_k W_k^*$ to build cache dist $\tilde{\mathbb{Q}}_k$, and calculate cache OT distance $d_{cOT}(\mathbb{P}_t, \tilde{\mathbb{Q}}_k; \tilde{C}_{t,k})$ 
           \STATE \textbf{Else} $d_{cOT} = 0$.
       \ENDFOR
       
       \STATE \textbf{Output:}  Predicted label $\hat{y} = \argmin\limits_{k \in [K]} \big( d_{OT}(\mathbb{P}_t, \mathbb{Q}_k; C_{t,k}) + \alpha d_{cOT}(\mathbb{P}_t, \tilde{\mathbb{Q}}_k; \tilde{C}_{t,k}) \big)$.
   \ENDFOR
\end{algorithmic}
\end{algorithm}